\documentclass{article}

    \PassOptionsToPackage{numbers, sort, compress}{natbib}

    \usepackage[final]{neurips_2022}

\usepackage[utf8]{inputenc} 
\usepackage[T1]{fontenc}    
\usepackage{url}            
\usepackage{booktabs}       
\usepackage{amsfonts}       
\usepackage{nicefrac}       
\usepackage{microtype}      
\usepackage[dvipsnames]{xcolor}         
\usepackage[colorlinks=true, citecolor=RoyalBlue]{hyperref}  
\usepackage{graphicx}
\usepackage{amsmath,amssymb} %
\usepackage{subfig}
\usepackage{enumitem}
\usepackage[font=small]{caption}
\usepackage{wrapfig}
\usepackage{pifont}
\newcommand{\cmark}{\ding{51}}
\newcommand{\xmark}{\ding{55}}

\newcommand{\revised}[1]{\textcolor{blue}{#1}}

\usepackage{cleveref}
\crefformat{section}{\S#2#1#3} 
\crefformat{subsection}{\S#2#1#3}
\crefformat{subsubsection}{\S#2#1#3}

\usepackage{xspace}

\makeatletter
\DeclareRobustCommand\onedot{\futurelet\@let@token\@onedot}
\def\@onedot{\ifx\@let@token.\else.\null\fi\xspace}

\def\ie{\emph{i.e}\onedot}

\makeatother

\usepackage{algorithm}
\usepackage{algorithmic}
\usepackage{bbding}
\usepackage{colortbl}
\usepackage{listings}
\usepackage{xspace}
\usepackage{multirow}
\usepackage{amssymb}
\usepackage{etoolbox}
\usepackage{color, colortbl}

\makeatletter
\renewcommand*{\ALG@name}{Architecture}
\makeatother

\newlength\savewidth
\newcommand{\tablestyle}[2]{\setlength{\tabcolsep}{#1}\renewcommand{\arraystretch}{#2}\centering\small}

\makeatletter\renewcommand\paragraph{\@startsection{paragraph}{4}{\z@}
  {.5em \@plus1ex \@minus.2ex}{-.5em}{\normalfont\normalsize\bfseries}}\makeatother

\title{Learning a Condensed Frame for Memory-Efficient Video Class-Incremental Learning}

\author{
  Yixuan Pei$^{1*}$ \quad Zhiwu Qing$^{2}$\thanks{equal contribution} \quad Jun Cen$^{3}$ \quad Xiang Wang$^{2}$ \quad Shiwei Zhang$^{4}$ \\ \textbf{Yaxiong Wang}$^{1}$ \quad \textbf{Mingqian Tang}$^{4}$ \quad \textbf{Nong Sang}$^{2}$ \quad \textbf{Xueming Qian}$^{1}$\\
  \\
  Xi'an Jiaotong University$^1$ \\
  Huazhong University of Science and Technology$^2$ \\
  The Hong Kong University of Science and Technology$^3$\\
  Alibaba Group$^4$\\
  \texttt{\{peiyixuan@stu, wangyx15@stu, qianxm@mail\}.xjtu.edu.cn, }\\
  \texttt{\{qzw,wxiang,nsang\}@hust.edu.cn,} \\
  \texttt{jcenaa@connect.ust.hk, \{zhangjin.zsw, mingqian.tmq\}@alibab-inc.com}\\
}

\begin{document}

\maketitle

\begin{abstract}
  Recent incremental learning for action recognition usually  stores representative videos to mitigate catastrophic forgetting.
  %
  However, only a few bulky videos can be stored due to the limited memory. 
  %
  %
  To address this problem, we propose \textit{FrameMaker}, a memory-efficient video class-incremental learning approach that learns to produce a condensed frame for each selected video. 
  %
  %
  Specifically, FrameMaker is mainly composed of two crucial components: \emph{Frame Condensing} and \emph{Instance-Specific Prompt}.
  The former is to reduce the memory cost by preserving only one condensed frame instead of the whole video, while the latter aims to compensate the lost spatio-temporal details in the Frame Condensing stage.
  By this means, FrameMaker enables a remarkable reduction in memory but keep enough information that can be applied to following incremental tasks.
 %
  %
  Experimental results on multiple challenging benchmarks, \ie, HMDB51, UCF101 and Something-Something V2, demonstrate that FrameMaker can achieve better performance to recent advanced methods while consuming only 20\% memory.
  %
  Additionally, under the same memory consumption conditions, FrameMaker significantly outperforms existing state-of-the-arts by a convincing margin.
  %
\end{abstract}

\section{Introduction}
Training video action recognition models with all classes in a single fine-tuning stage have been widely studied~\cite{carreira2017i3d,lin2019tsm,tran2015c3d,xie2017s3d,tran2018r21d,li2022uniformer,fan2021mvit,liu2021videoswin} in recent years.
%
%
However, in many realistic scenarios, limited by privacy matters or technologies, the different classes can only be presented in sequence, where the previously trained classes are either unavailable or partially available in limited memory.
Naively training models will overfit currently available data and cause performance deterioration on previously seen classes, which is named catastrophic forgetting~\cite{mccloskey1989catastrophic}.
Class-incremental learning is a  machine learning paradigm to fight this challenge when fine-tuning a single model in a sequence of independent classes.
%


A branch of class-incremental learning methods~\cite{castro2018e2eil,dhar2019lwm,douillard2020podnet,hou2019ucir,rebuffi2017icarl,li2017lwfmc,wu2019bic,buzzega2020glc} has achieved remarkable performances in the image domain by re-training a portion of past examples. 
%
Meanwhile, some existing video incremental learning methods~\cite{park2021tcd,villa2022vclimb} have demonstrated that storing more past examples can improve the ability to mitigate forgetting.
%
Nevertheless, videos are information-redundant and require a high memory load to be saved, hence it is impractical to store a good deal of training videos for each class. 
%
Therefore, herding strategy~\cite{rebuffi2017icarl} is adopted to select a small set of representative videos for the exemplar memory~\cite{park2021tcd,villa2022vclimb}.
%
vCLIMB~\cite{villa2022vclimb} tries to down-sample frames for each video to reduce the memory consumption, and imposes a temporal-consistency regularization for better performance.
%
Despite the remarkable performance, these methods still demand to conserve multiple (\textit{i.e.}, 8-16) frames for each representative video, leading to non-negligible memory overhead, which limits their further real-world applications.

  
\begin{figure}
    \centering
    \includegraphics[width=1.0\linewidth]{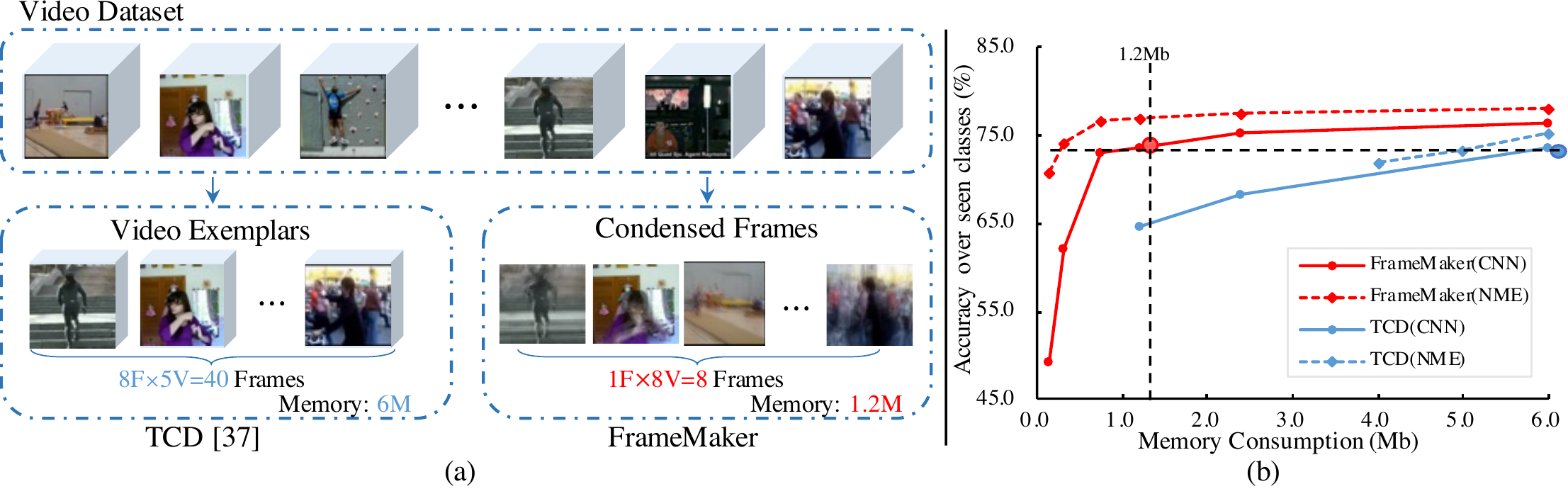}
    \caption{An intuitive comparison between FrameMaker and TCD~\cite{park2021tcd}. 
    (a) Compared with TCD, FrameMaker stores only one condensed frame for each video exemplar. 
    (b) Illustration of memory consumption and accuracy curve. 
    FrameMaker costs 1.2Mb memory, which slightly exceeds the performance of TCD with 6.0Mb memory.
    }
    \label{fig:intuitive_comparision}
\end{figure}


To remedy the above weaknesses, we present FrameMaker, a memory-efficient video class-incremental learning approach.
%
%
Concretely, the proposed FrameMaker contains two components, Frame Condensing and Instance-Specific Prompt.
In Frame Condensing, FrameMaker assigns a learnable weight to each frame, and then these weights are adjusted to make the weighted sum frame \ie, the condensed frame, share the same embedding features with the original video clip.
Obviously, the collapsed temporal dimension and mixed spatial cues by Frame Condensing may degrade the accuracy of action recognition.
%
%
To compensate the missing spatio-temporal details, inspired by Task-Specific Prompt~\cite{liu2021prompt}, we further propose Instance-Specific Prompt, which learns a set of parameters for each condensed frame to perform fine-grained pixel-wise modifications.
%
In this way, the memory required to store a single video can be remarkably compressed.
An intuitive comparison is given in Figure~\ref{fig:intuitive_comparision}(a). 
On the premise of comparable performance with TCD~\cite{park2021tcd}, 
FrameMaker significantly reduces memory overhead by 80\%.
%
Moreover, the comparison curve displayed in Figure~\ref{fig:intuitive_comparision}(b) further indicates that FrameMaker achieves a consistent performance improvement over recent TCD with the same memory consumption.
%
%
%

In summary, we make the following contributions: 
1) We present a memory-efficient video class-incremental learning method FrameMaker, which significantly reduces the memory consumption of example videos by integrating multiple frames into one condensed frame; 
2) For the first time, we show a novel perspective of Prompt, \textit{i.e.,} Instance-Specific Prompt, and demonstrate that it can embed effective spatio-temporal interactions into learned condensed frames for video class-incremental learning;
3) Experimental results reveal that FrameMaker consistently outperforms recent state-of-the-art methods with the same memory overhead on three challenging benchmarks (HMDB51, UCF101, and Something-Something V2).


\section{Related Works}

\subsection{Class-Incremental Learning}
The class-incremental learning has been well-studied in \textbf{image domain}~\cite{aljundi2018mas,li2017img_reg4,kirkpatrick2017img_reg2,zenke2017img_reg3}, existing approaches could be grouped into knowledge distillation-based methods ~\cite{hinton2015distilling}, memory-based methods~\cite{ostapenko2019gen1,shin2017gen2,rebuffi2017icarl}, and architecture-based methods.  
\textbf{In video domain}, class-incremental learning for action recognition is still an underexplored area. 
The existing solutions are designed for better temporal constraints, such as decomposing the spatio-temporal features~\cite{zhao2021civc}, exploiting time-channel importance maps~\cite{park2021tcd}, and applying temporal consistency regularization~\cite{villa2022vclimb}.
Notably, both ~\cite{park2021tcd} and ~\cite{villa2022vclimb} have shown that more stored examples in memory can effectively encourage the performance.
However, FrameMaker attempts to minimize the stored frames for each video by a pretty maneuver, which can provide decent performance with slight storage requirements.

\subsection{Prompt-Based Learning}
Prompting is an emerging transfer learning technique in natural language processing~\cite{liu2021prompt}, which adapts the language model to downstream tasks by a fixed function.
Since the design of prompting functions is challenging, the learnable prompts~\cite{lester2021learnable_prompt1,li2021learnable_prompt2} are proposed to learn task-specific knowledge parameters for the input.
Recently, in the visual domain, some works have also emerged to adapt the pre-trained models to new tasks by adding prompt tokens~\cite{jia2022visual_prompt_token} or pixels~\cite{bahng2022visual_prompt_pixel} in the data space. 
Besides, DualPrompt~\cite{wang2022dualprompt} and L2P~\cite{wang2021l2p} get rid of the rehearsal buffer in continual learning by learning a set of task-specific parameters.
In this paper, we draw inspiration from prompting, and present an Instance-Specific Prompt to replenish the collapsed spatio-temporal cues for condensed frames.

\subsection{Action Recognition}
The approaches designed for action recognition task can be grouped into three categories: 2D CNN, 3D CNN and Transformer-based methods. 2D CNN based methods is more efficient~\cite{he2016resnet,simonyan2014vgg}, while 3D CNN~\cite{feichtenhofer2019slowfast,carreira2017i3d,feichtenhofer2020x3d,tran2015c3d,tran2019csn,diba2018stc,xie2017s3d,qiu2017p3d,wang2018atrnet,tran2018r21d} achieves better accuracy in action recognition at the cost of more computation cost. With the rapid development of vision transformer~\cite{liu2022learning,hu2019local,liu2021swinv2}, researchers also apply it into the video recognition~\cite{arnab2021vivit,bertasius2021timesformer,liu2021videoswin,fan2021mvit}, which is proven to be remarkable effective.
%
%
%
In this work, we follow the settings in ~\cite{park2021tcd,zhao2021civc}, and employ the Temporal Shift Module (TSM)~\cite{lin2019tsm} based on 2D CNN for efficient experiments.

The video summarization and efficient action recognition tasks have provided some methods for key-frame selection and generation, which are related to this work. 
The video summarization~\cite{zhu2020dsnet, wang2019stacked,zhao2021reconstructive} task selects the most representative frames for efficient human understanding, which has a different paradigm with action recognition task, hence we will mainly discuss efficient 
The efficient video classification task aims to learn a strategy to improve classification accuracy with fewer discriminative frames as input. The approaches designed for this task can be grouped into two categories: rule-based and model-based methods. 
The former~\cite{korbar2019scsampler, zhi2021mgsampler} select frames via some pre-defined rules such as motion distribution~\cite{zhi2021mgsampler} or channel sampling~\cite{kim2022greyst}, and the later design a submodel to select or generate key frames~\cite{qiu2021condensing, huang2018makes, wu2019adaframe, wu2019multi, fan2018watching, gowda2021smart}. 
In this paper, we propose model-free frame condensing method for video class-incremental learning, which only learns a few condensing weights and prompting parameters for each instance.

\begin{figure}
  \centering
  \includegraphics[width=1.0\linewidth]{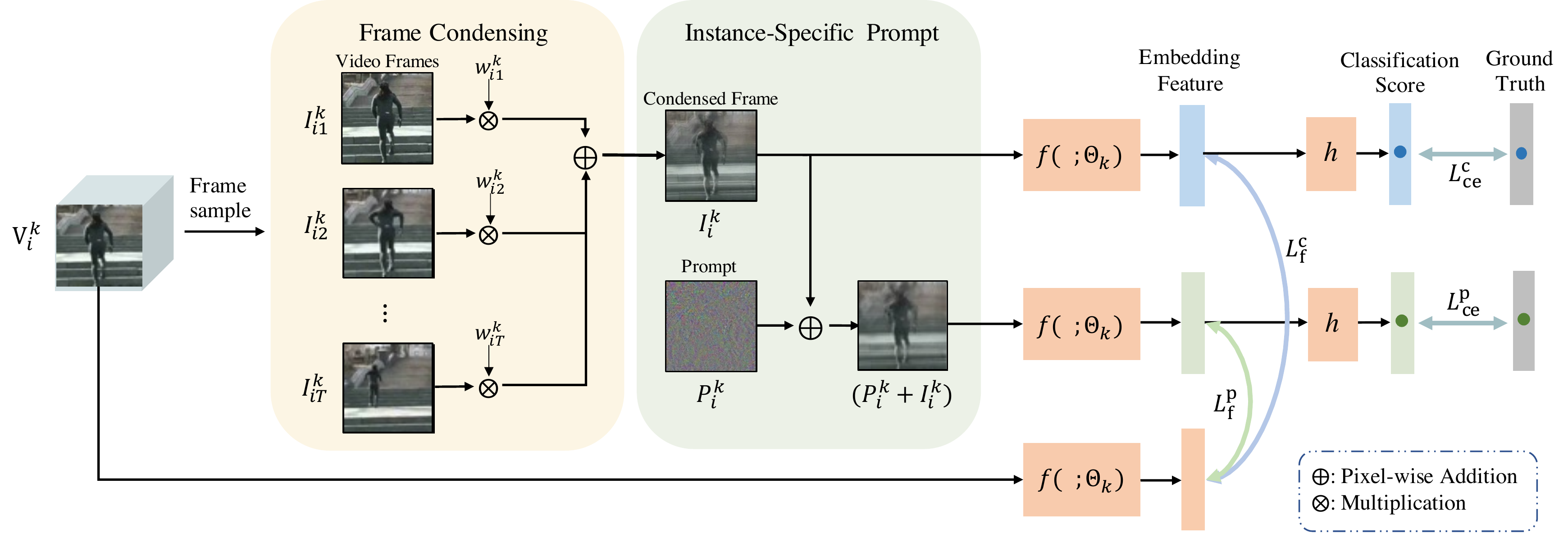}
  \caption{An overview of the proposed FrameMaker, which mainly contains Frame Condensing and Instance-Specific Prompt. 
  For each selected representative video $V_i^k$, we first uniformly sample $T$ frames and learn a condensing weight $w_i^k$ for each frame to perform Frame Condensing. Then a learnable prompt $P_i^k$ is introduced for the condensed frame to compensate the lost spatio-temporal information. Finally, distillation loss and cross entropy loss are both used to guide the optimization of condensing weights and prompt.
  }
  \label{fig:overview}
\end{figure}

\section{Method}

This section presents the general framework of FrameMaker as shown in Figure~\ref{fig:overview}, which mainly consists of two crucial components: Frame Condensing and Instance-Specific Prompt.
Firstly, Frame Condensing is designed to integrate multiple frames into one frame for efficient storage.
Then we introduce Instance-Specific Prompt to compensate the collapsed temporal information and the mixed spatial knowledge.
Finally, the specific class-incremental training details for subsequent tasks using Condensed Frames are given.

\subsection{Problem formulation}
The purpose of video class-incremental learning is to train a model $F(;\Theta)=h(f(;\theta);\xi)$ parameterized by $\Theta$ as the tasks $\{\mathcal{T}^1,\mathcal{T}^2,\cdots,\mathcal{T}^K\}$ arrive in a sequence, where $f(;\theta)$ is the feature extractor and $h(;\xi)$ is the classification head. 
Each task $\mathcal{T}^K$ has its specific dataset $D^k=\{(d_i^k,y_i^k),y_i^k\in{L^k}\}$ whose labels belong to a predefined label set $L^k$ and have not appeared in previous tasks. 
The memory-based methods~\cite{rebuffi2017icarl,villa2022vclimb,park2021tcd} have been shown to be effective in preventing catastrophic forgetting.
Specifically, after the incremental training step $k$, a memory bank $M^k$ will be established for the dataset $D^k$ in this task to store the representative exemplars or features. Exemplars stored in previous datasets $M^{1:k}=M^1\bigcup{M^2}\bigcup{}\cdots\bigcup{M^k}$ will be used in subsequent incremental tasks to mitigate forgetting. At the incremental step $k$, the model $F(;\Theta_k)$ is trained from $F(;\Theta_{k-1})$ with ${D'}^k=D^k\bigcup{M^{1:(k-1)}}$ and evaluated on all seen classes.

\subsection{FrameMaker}
\paragraph{Frame Condensing.}

We follow the standard protocol of video class-incremental methods~\cite{villa2022vclimb,park2021tcd}, which are based on memory-replay strategy and knowledge distillation.
Differently, we devotes to reducing the number of stored frames for representative videos to achieve memory-efficient video class-incremental learning.
Formally, we first select a subset of video instances $V^k$ by herding strategy~\cite{rebuffi2017icarl} from $D^k$ after the incremental step $k$.
%
%
Given a video $V_i^k\in{V^k}$, we then sample $T$ frames $\{I_{i1}^k,I_{i1}^k,\cdots,I_{iT}^k\}$ uniformly and integrate them to condense a more representative one. 
%
To this end, we define learnable weights $W_i^k=\{w_{i1}^k,w_{i2}^k,\cdots,w_{iT}^k\}$ for the $i$-th video instance $V_i^k$ in exemplar set $V^k$. And the condensed frame can be calculated as follows:
\begin{equation}
    I_i^k=\sum_{t=1}^{T}\frac{e^{w_{it}^k}}{\sum_{t=1}^{T}e^{w_{it}^k}}{I_{it}^k}\in{\mathbb{R}^{{C}\times{{H}}\times{{W}}}} \ ,
    \label{eq:frame_condensing}
\end{equation}
where $C$, $H$ and $W$ are the size of channel, height and width of input frames, respectively.
Next, the condensed frame $I_i^k$ is excepted to empower the same or much similar expressive ability as the original video clip.
%
Hence, the embedding features of the condensed frame extracted from current model should be consistent with the features of the original video clip:
\begin{equation}
    L^\text{c}_{\text{f}}={\mid\mid{f(I_i^k;\theta_k)-f(V_i^k;\theta_k)}\mid\mid}^2 \ ,
    \label{eq:feat_consis1}
\end{equation}
where $f(;\theta_k)$ is the feature extractor of current model $F(;\Theta_k)$. This consistency regularization can foster the condensing weights to exploit memory-efficient expressions of video clips. 
To further improve the adaptability of the condensed frames to the current model, we employ cross entropy loss to supervise the classification confidence from the condensed frame:
\begin{equation}
    L^\text{c}_{\text{ce}}={\text{CrossEntropy}(F(I_i^k;\Theta_k),y_i^k)} \ .
\end{equation}
The full objective function for condensing weights $W_i^k$ is given by:
\begin{equation}
    L_{\text{c}}={{L^\text{c}_{\text{f}}}+L^\text{c}_{\text{ce}}}  \ .
\end{equation}
With the methods described above, we can learn an effective representation, \emph{i.e.}, the condensed frame for each video based on current model.

\paragraph{Instance-Specific Prompt.}
However, 
condensing a video into one frame would inevitably lose the the temporal dynamics and complete spatial information of the original video to a certain extent.
Therefore, an intriguing perspective of prompt, termed as Instance-Specific Prompt, is proposed to replenish the missing spatio-temporal cues for the condensed frame.

Specifically, we first construct a learnable \textit{prompt} $P_i^k\in{\mathbb{R}^{{C}\times{{H}}\times{{W}}}}$ for each exemplar video $V_i^k$, which shares the same spatial resolution as its original clip. 
%
The prompt $P_i^k$ is then pixel-wise summed with the condensed frame $I_i^k$, and learns to pull the embedding feature of the summed frame $(I_i^k+P_i^k)$ and the video clip $V_i^k$ together, which is similar to Eq.~\ref{eq:feat_consis1}:
\begin{equation}
    L^\text{p}_{\text{f}}={\mid\mid{f((I_i^k+P_i^k);\theta_k)-f(V_i^k;\theta_k)}\mid\mid}^2 \ .
    \label{eq:feat_consis2}
\end{equation}
Notably, the Eq.~\ref{eq:feat_consis1} for condensing weights is difficult to learn the satisfactory spatio-temporal features due to the collapse of the temporal dimension. 
Nevertheless, the Eq.~\ref{eq:feat_consis2} can enrich the representations of condensing frames by introducing more flexible learnable parameters in the input space.
Besides, the Cross-Entropy loss is also utilized to enhance the semantic perception:
\begin{equation}
    L^\text{p}_{\text{ce}}=\text{CrossEntropy}(F((I_i^k+P_i^k);\Theta_k),y_i^k) \ .
\end{equation}
Theoretically, the condensing weights $W_i^k$ and the prompting parameters $P_i^k$  can be jointly updated by $L_\text{p}={L^\text{p}_{\text{f}} + L^\text{p}_{\text{ce}}}$.
Interestingly, we observed in our practice that the flexible prompt $P_i^k$ leads to the condensing weights $W_i^k$ being under optimized.
Hence, we give the final training objective function for frame condensing as:
\begin{equation}
    L_{\text{fc}}={\alpha}L^\text{c}_{{\text{f}}} + {\beta}L^\text{c}_{{\text{ce}}} + {\gamma}L^\text{p}_{{\text{f}}} + {\eta}L^\text{p}_{\text{ce}} \ ,
\end{equation}
where additional $L^\text{c}_{\text{f}}$ and $L^\text{c}_{\text{ce}}$ are added to achieve stronger constraint on $W_i^k$ for its effective training. ${\alpha}$, ${\beta}$, ${\gamma}$ and ${\eta}$ are the balance weights for each term, and they are empirically set to 1.0 unless otherwise specified.

After the optimization of condensing weights and prompting parameters, the prompt is added directly to the condensed frame and stored together. And the learned
prompts will be frozen when the corresponding task is completed. Therefore, there is no extra memory cost for prompts. 

\subsection{Training}
It is worth noting that FrameMaker aims to condense the stored frames for representative videos, and other class-incremental learning steps still follow the standard pipeline.
Specifically, when training the incremental step $k$, we use the dataset ${D'}^k=D^k\bigcup{M^{k-1}}$ to update the model from $F(;\Theta_{k-1})$,
where $D^k$ is the dataset of the task $k$ which consists of videos belonging to ${L^k}$, and ${M^{k-1}}$ is memory bank which contains condensed frames generated by FrameMaker.
The training samples from $D^k$ and ${M^{k-1}}$ are alternately fed into the current model according to the proportion of their sample number.
%
To further prevent the catastrophic forgetting, we also employ the knowledge distillation method proposed in PODNet~\cite{douillard2020podnet} to transfer the knowledge from the previous model $F(;\Theta_{k-1})$ to current model $F(;\Theta_{k})$.
The overall objective function for task $k$ reads:
\begin{equation}
    L_{\text{cil}}={L^\text{d}_{{\text{ce}}} + L^\text{m}_{{\text{ce}}} + L^\text{m}_{{\text{dist}}}} \ ,
\end{equation}
where $L^\text{d}_{{\text{ce}}}$ and $L^\text{m}_{{\text{ce}}}$ are the Cross Entropy losses for new task data $D^k$ and condensed frame exemplars ${M^{k-1}}$, respectively. And $L^\text{m}_{{\text{dist}}}$ is the knowledge distillation loss~\cite{douillard2020podnet} for condensed frame exemplars. 


\section{Experiments}
\subsection{Experimental Setup}
\paragraph{Datasets and evaluation metrics.}
The proposed FrameMaker is evaluated on three standard action recognition datasets, UCF101~\cite{soomro2012ucf101}, HMDB51~\cite{kuehne2011hmdb} and Something-Something V2~\cite{goyal2017something}. HMDB51 dataset consists of 6.8K videos belonging to 51 classes from YouTube or other websites. UCF101 dataset contains 13.3K videos from 101 classes. Something-Something V2 is a crowd-sourced dataset that includes 220K videos from 174 classes. 

For UCF101, the model is trained on 51 classes first, and the remaining 50 classes are divided into 5, 10 and 25 tasks. For HMDB51, we train the base model using videos from 26 classes, and the remaining 25 classes are separated into 5 or 25 groups. For Something-Something V2, we first train 84 classes in the initial stage, and generate the groups of 10 and 5 classes.

To evaluate the performance of class-incremental learning, we infer the test videos from all seen categories after each task, and finally, report the average accuracy of all tasks. 
Following ~\cite{park2021tcd}, two different metrics are reported, \textit{i.e.,} CNN and NME. 
CNN refers to training a fully-connected layer for extracted features, which is a standard classification protocol.
NME is proposed by iCaRL~\cite{rebuffi2017icarl}, which assigns the labels for test data by comparing the feature embeddings with the mean-of-exemplars.

\paragraph{Implementation details.}
TSM~\cite{lin2019tsm} is employed as our backbone, and we follow the data preprocessing procedure of TSM. 
For UCF101, we train a ResNet-34 TSM for 50 epochs with a batch size 256 from an initial learning rate 0.04. For HMDB51 and Something-Something V2, we train a ResNet-50 TSM for 50 epochs with a batch size of 128 from an initial learning rate of 1e-3 and 0.04, respectively. 
All used networks are first pre-trained on ImageNet~\cite{deng2009imagenet} for initialization.
These settings are consistent with TCD~\cite{park2021tcd}.
We train all models on eight NVIDIA V100 GPUs and use PyTorch~\cite{paszke2019pytorch} for all our experiments.

\begin{table}
  \caption{Comparison with the state-of-the-art approaches over class-incremental action recognition performance on UCF101 and HMDB51. Our FrameMaker achieves the best performance under all experimental settings.}
  \label{table:sota_ucf_hmdb}
  \tablestyle{6pt}{1}
  \centering
  \begin{tabular}{ccccccccccc}
    \toprule[1.15pt]
    ~ &  \multicolumn{6}{c}{UCF101} & \multicolumn{4}{c}{HMDB51}   \\
    \cmidrule(r){1-1}\cmidrule(r){2-7}\cmidrule(r){8-11}
    Num. of Classes  & \multicolumn{2}{c}{$10\times5$ stages}   & \multicolumn{2}{c}{$5\times10$ stages}   & \multicolumn{2}{c}{$2\times25$ stages}  & \multicolumn{2}{c}{$5\times5$ stages}  & \multicolumn{2}{c}{$1\times25$ stages}\\
    Classifier  & CNN & NME & CNN & NME & CNN & NME & CNN & NME & CNN & NME   \\
    \cmidrule(r){1-1}\cmidrule(r){2-3}\cmidrule(r){4-5}\cmidrule(r){6-7}\cmidrule(r){8-9}\cmidrule(r){10-11}
    Finetuning & 24.97  & - & 13.45 & - & 5.78 & - & 16.82 & - & 4.83 & -     \\
    LwFMC~\cite{li2017lwfmc}   & 42.14 & - & 25.59 & - & 11.68 & - & 26.82 & - & 16.49 & -   \\
    LwM~\cite{dhar2019lwm}     & 43.39 & - & 26.07 & - & 12.08 & - & 26.97 & - & 16.50 & -  \\
    iCaRL~\cite{rebuffi2017icarl}   & - & 65.34 & - & 64.51 & - & 58.73 & - & 40.09 & - & 33.77    \\
    UCIR~\cite{hou2019ucir}    & 74.31 & 74.09 & 70.42 & 70.50 & 63.22 & 64.00 & 44.90 & 46.53 & 37.04 & 37.15   \\
    PODNet~\cite{douillard2020podnet}  & 73.26 & 74.37 & 71.58 & 73.75 & 70.28 & 71.87 & 44.32 & 48.78 & 38.76 & 46.62  \\
    TCD~\cite{park2021tcd}     & 74.89 & 77.16 & 73.43 & 75.35 & 72.19 & 74.01 & 45.34 & 50.36 & 40.47 & 46.66  \\
    \cmidrule(r){1-1}\cmidrule(r){2-3}\cmidrule(r){4-5}\cmidrule(r){6-7}\cmidrule(r){8-9}\cmidrule(r){10-11}
    FrameMaker  & \textbf{78.13} & \textbf{78.64} & \textbf{76.38} & \textbf{78.14} & \textbf{75.77} & \textbf{77.49} & \textbf{47.54} & \textbf{51.12} & \textbf{42.65} & \textbf{47.37} \\
    \bottomrule[1.15pt]
  \end{tabular}
\end{table}

\begin{table}[]
\begin{minipage}[t]{0.46\textwidth}
\centering
\caption{{Comparison with the top approaches over class-incremental action recognition performance on Something-Something V2.}}
\label{table:sota_ssv2}
\tablestyle{4pt}{1.5}
\begin{tabular}{ccccc}
    \toprule[1.15pt]
     Num. of Classes & \multicolumn{2}{c}{$10\times9$ stages}   & \multicolumn{2}{c}{$5\times18$ stages}  \\
    Classifier  & CNN & NME & CNN & NME   \\
    \cmidrule(r){1-1}\cmidrule(r){2-3}\cmidrule(r){4-5}
    UCIR~\cite{hou2019ucir} & 26.84 & 17.98 & 20.69 & 12.57 \\
    PODNet~\cite{douillard2020podnet} & 34.94 & 27.33 & 26.95 & 17.49 \\
    TCD~\cite{park2021tcd} & 35.78 & 28.88 & 29.60 & 21.63  \\
    \cmidrule(r){1-1}\cmidrule(r){2-3}\cmidrule(r){4-5}
    FrameMaker & \textbf{37.25} & \textbf{29.92} & \textbf{30.98} & \textbf{22.84} \\
    \bottomrule[1.15pt]
  \end{tabular}
\end{minipage}
\hspace{1.5mm}
\begin{minipage}[t]{0.5\textwidth}
\centering
    \tablestyle{4pt}{1}
    \caption{{Ablations for Frame Condensing (FC) and Instance-Specific Prompting (ISP) on UCF101 with 10 steps and HMDB51 with 5 steps.}}
    \label{tab:fc_isp}
\begin{tabular}{ccccccc}
    \toprule[1.15pt]
    ~ & ~ & ~ & \multicolumn{2}{c}{UCF101} & \multicolumn{2}{c}{HMDB51}   \\
    \cmidrule(r){1-1}\cmidrule(r){2-3}\cmidrule(r){4-5}\cmidrule(r){6-7}
    Frames & FC & ISP & CNN & NME & CNN & NME \\
    \cmidrule(r){1-1}\cmidrule(r){2-3}\cmidrule(r){4-5}\cmidrule(r){6-7}
    All & - & - & 72.09 & 75.70 & 43.38 & \textbf{47.00}  \\
    Random& \xmark & \xmark & 68.64 & 73.96 & 39.59 & 43.48  \\
    Random& \xmark & \cmark & 70.71 & 75.04 & 39.81 & 43.74   \\
    Average& \xmark & \xmark & 70.82&75.45&41.84&45.45  \\
    Average& \xmark & \cmark & 71.51 &76.23 &42.59 &46.46   \\
    Condensed& \cmark & \xmark & 72.29 & 76.42 & 42.18 & 46.27 \\
    Condensed& \cmark & \cmark & \textbf{72.93} & \textbf{76.64} & \textbf{43.39} & 46.88  \\
    \bottomrule[1.15pt]
  \end{tabular}
\end{minipage}
\end{table}

\subsection{Comparison with State-of-the-art Results}
This section presents the qualitative comparison of our proposed FrameMaker with the existing class incremental learning approaches under multiple challenging settings on three datasets: UCF101~\cite{soomro2012ucf101}, HMDB51~\cite{kuehne2011hmdb} and Something-Something V2~\cite{goyal2017something}.
For a fair comparison, we use the same exemplar memory size per class, model structure and pre-training initialization as the existing methods~\cite{park2021tcd}.

Table~\ref{table:sota_ucf_hmdb} summarizes the results on UCF101 and HMDB51, which shows that FrameMaker outperforms other methods consistently under different configurations in terms of both CNN and NME scores. 
%
%
The average accuracy of our FrameMaker surpasses TCD by around 3.0\% on UCF101 and 2.0\% on HMDB51 in terms of CNN, respectively. 
These results demonstrate that one condensed frame can also be equipped with an effective spatio-temporal representation.
%
%

FrameMaker is compared with recent advanced methods on Something-Something V2 in Table~\ref{table:sota_ssv2}. 
FrameMaker sets new state-of-the-art performance under multiple challenging settings.
Although FrameMaker only uses one condensed frame for future replay, it still achieves decent performance on the large-scale motion-sensitive dataset.
We speculate that these gains mainly come from two aspects:
(1) The proposed Instance-Specific Prompt effectively rescues the lost spatio-temporal details, which is also discussed in Section~\ref{sec:ablation}.
(2) Memory-efficient FrameMaker allows us to store more exemplar videos with less memory consumption, which greatly alleviates catastrophic forgetting.

\subsection{Ablation Study}
\label{sec:ablation}
In this section, we present ablation studies to analyze the properties and the effectiveness of FrameMaker.
If not specified, the ablation studies are performed on UCF101 with 10 steps and HMDB51 with 5 steps. 
For comparison with the case of storing all frames, we select 5 exemplar videos for each class.

\begin{table}[]
\begin{minipage}[t]{0.30\textwidth}
\centering
\caption{Ablations for objective functions in Frame Condensing on UCF101 with 10 steps.}
\label{tab:loss_fc}
\tablestyle{7.5pt}{1}
\begin{tabular}{cccc}
    \toprule[1.15pt]
    ${L^\text{c}_{\text{f}}}$ & $L^\text{c}_{\text{ce}}$ & CNN & NME \\
    \cmidrule(r){1-2}\cmidrule(r){3-4}
    \xmark & \xmark & 70.82 & 75.45  \\
    \cmark & \xmark & 71.35 & 75.76  \\
    \xmark & \cmark & 71.96 & 76.07   \\
    \cmark & \cmark & \textbf{72.29} & \textbf{76.42} \\
    \bottomrule[1.15pt]
  \end{tabular}
\end{minipage}
\hspace{3mm}
\begin{minipage}[t]{0.355\textwidth}
\centering
\caption{Ablations for objective functions in Instance-Specific Prompt on UCF101 with 10 steps.}
\label{tab:loss_isp}
\tablestyle{7.5pt}{1}
\begin{tabular}{cccccc}
    \toprule[1.15pt]
    $L^\text{c}_{\text{*}}$ & $L^\text{p}_{\text{f}}$ & $L^\text{p}_{^\text{ce}}$ & CNN & NME \\
    \cmidrule(r){1-3}\cmidrule(r){4-5}
    \xmark & \cmark & \cmark & 71.83 & 75.94   \\
    \cmark & \cmark & \xmark & 72.35 & 76.48   \\
    \cmark & \xmark & \cmark & 72.58 & 76.57  \\
    \cmark & \cmark & \cmark & \textbf{72.93} & \textbf{76.64}  \\
    \bottomrule[1.15pt]
  \end{tabular}
\end{minipage}
\hspace{4mm}
\begin{minipage}[t]{0.265\textwidth}
\centering
  \caption{The number of frames for Frame Condensing on UCF101 with 10 steps.}
  \label{tab:fram_num}
  \centering
  \tablestyle{7.5pt}{1}
  \begin{tabular}{ccc}
    \toprule[1.15pt]
    $T$  & CNN & NME \\
    \cmidrule(r){1-1}\cmidrule(r){2-3}
    2 & 71.62 & 75.89  \\
    4 & 71.70 & 76.12  \\
    8 & \textbf{72.93} & \textbf{76.64}  \\
    16 & 72.42 & 76.11  \\
    \bottomrule[1.15pt]
  \end{tabular}
\end{minipage}
\end{table}

{\color{blue}
\begin{table}[]
\begin{minipage}[t]{0.45\textwidth}
    \centering
    \caption{{Ablations for different backbones for FrameMaker on HMDB51 with 5 steps.}}
    \tablestyle{5pt}{1.0}
    \begin{tabular}{ccccc}
        \toprule[1.15pt]
        Backbone & Frames & CNN & NME & Mem. \\
        \cmidrule(r){1-2}\cmidrule(r){3-4}\cmidrule(r){5-5}
        TSM & All & 43.38 & \textbf{47.00} & 6.00Mb\\
        TSM & FC+ISP &  \textbf{43.39} & 46.88 & 0.75Mb \\ 
        \cmidrule(r){1-2}\cmidrule(r){3-4}\cmidrule(r){5-5}
        R3D50 & All  & 39.85 & \textbf{45.64} & 6.00Mb \\
        R3D50 & FC+ISP & \textbf{39.88} & 45.08 & 0.75Mb\\
        \cmidrule(r){1-2}\cmidrule(r){3-4}\cmidrule(r){5-5}
        ViT & All  & \textbf{35.34} & 39.46 & 6.00Mb \\
        ViT & FC+ISP& 35.25 & \textbf{39.58} & 0.75Mb\\
        \bottomrule[1.15pt]
    \end{tabular}
    \label{tab:backbone}
\end{minipage}
\hspace{4mm}
\begin{minipage}[t]{0.475\textwidth}
    \centering
    \caption{{Ablations for alternative prompting strategies for FrameMaker on HMDB51 with 5 steps.}
    }
    \tablestyle{5pt}{1.07}
    \begin{tabular}{ccccc}
        \toprule[1.15pt]
        Position & Type & CNN & NME & Mem. \\
        \cmidrule(r){1-2}\cmidrule(r){3-4}\cmidrule(r){5-5}
        Feature & T.-Spec.  & 41.67  &  45.71  &  77.07Mb \\ 
        Feature & C.-Spec.  & 41.72  &  45.93  &  385.35Mb \\ 
        Feature & I.-Spec.  & 42.19 &  46.53  & 1926.75Mb \\ 
        \cmidrule(r){1-2}\cmidrule(r){3-4}\cmidrule(r){5-5}
        Frame & T.-Spec.  & 41.16 &  45.34  & 0.75Mb \\ 
        Frame & C.-Spec.  & 42.01 & 45.47  & 0.75Mb \\ 
        Frame & I.-Spec. & \textbf{43.39} & \textbf{46.88} & 0.75Mb \\ 
        \bottomrule[1.15pt]
    \end{tabular}
    \label{tab:prompt_strategies}
\end{minipage}
\end{table}
}

\paragraph{Frame Condensing and Instance-Specific Prompt.} To show the effectiveness of Frame Condensing and Instance-Specific Prompt,  we compare them with the following alternative methods to produce stored frames for future replay:
(1) \textit{All} Frames, which simply saves the whole video.
(2) One \textit{Random} Frame, which randomly selects one frame for each exemplar video.
(3) One \textit{Average} Frame, which averages the input frames for each exemplar video.
(4) One \textit{Condensed} Frame, which is generated by our Frame Condensing procedure.
The results are reported in Table~\ref{tab:fc_isp}. 
From the table, in terms of CNN, we observe that Instance-Specific Prompt yields an improvement of 2.07\% and 0.22\% with randomly sampled frames, 0.69\% and 0.75\% with averaged frames on UCF101 and HMDB51, respectively,  which implies that prompts can always replenish the ignored spatio-temporal cues in video frames.
Further, Frame Condensing with Instance-Specific Prompt leads to at least on par or better performance than preserving all frames while using just one condensed frame under all settings on both datasets.
%
%
%
%

\begin{figure}[t]
  \centering
  \includegraphics[width=1.0\linewidth]{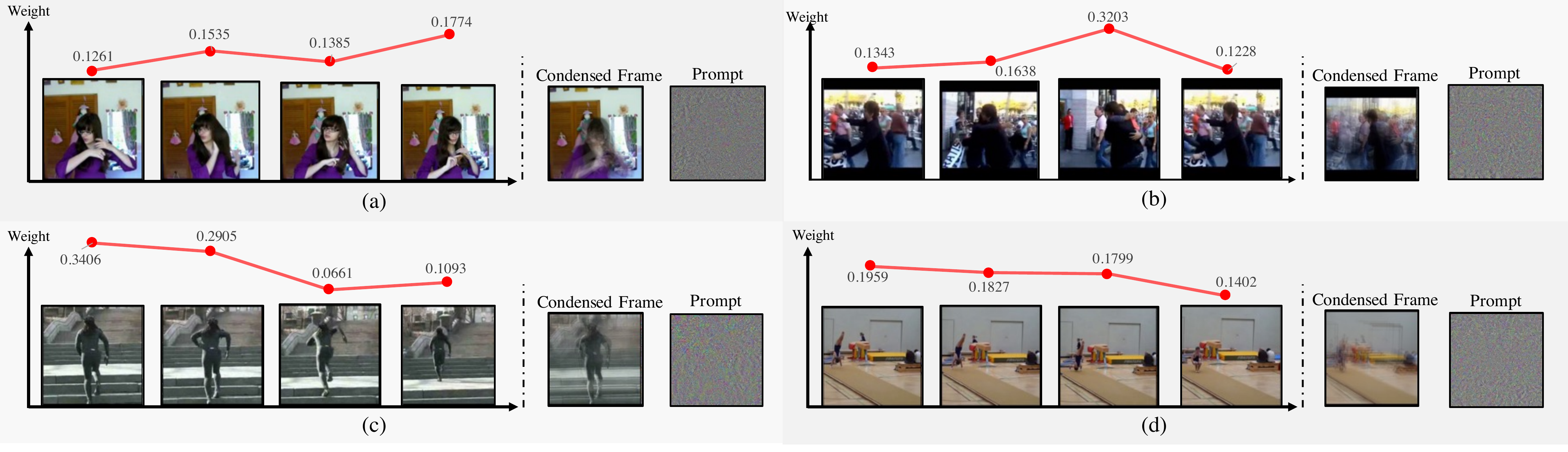}
  \caption{Visualization of Condensing Weights and Instance-Specific Prompts on HMDB51.}
  \label{fig:vis_weight_prompt}
\end{figure}

\paragraph{Loss terms.}
To preserve the best spatio-temporal information in the condensed frames, the distillation loss $L^\text{*}_\text{f}$ and the cross entropy loss $L^\text{*}_\text{ce}$ are both applied for Frame Condensing and Instance-Specific Prompt.
%
%
Table~\ref{tab:loss_fc} presents the results with Frame Condensing from several different combinations of loss terms. 
%
%
%
$L^\text{c}_\text{ce}$ provides more explicit semantic supervision, making the accuracy to step further.
%
%
%
Similar experiments for Instance-Specific Prompt are shown in Table~\ref{tab:loss_isp}. 
One noticeable thing is that once introducing prompt, we only need to calculate the final $L^\text{p}_\text{f}$ and $L^\text{p}_\text{ce}$ to realize the joint update of condensing weights $W^k_i$ and prompt $P^k_i$.
However, we observe that the learned weights, in this case, tend to be average, which indicates that $W^k_i$ is under-optimized. %
We speculate that this is due to the strong plasticity of prompt, which leads to the insufficient optimization of $W^k_i$.
To this end, we further employ $L^\text{c}_\text{f}$ and $L^\text{c}_\text{ce}$ for $W^k_i$ as the guidance.
The improvement in Table~\ref{tab:loss_isp} and the visualizations in Figure~\ref{fig:vis_weight_prompt} show that $W^k_i$ receives effective learning.

\paragraph{The number of frames used for Frame Condensing.}
As discussed in Section 3.2, the $T$ frames are uniformly sampled from a representative video for Frame Condensing.
In Table~\ref{tab:fram_num}, we show the effect of $T$ in video class-incremental learning.
We observe that the performance increases as $T$ increases for both CNN and NME.
However, the performance is saturated when using 16 frames, which is in line with the conclusion of existing methods~\cite{villa2022vclimb,park2021tcd}, \textit{i.e.,} storing more frames in a video does not necessarily deliver performance improvement.
This phenomenon reveals that FrameMaker can empower one Condensed Frame to absorb plenty of spatio-temporal features from original clips.


\paragraph{Different backbones.} To evaluate the applicability of proposed FrameMaker, we replace {the backbone with 3D CNN-based method R3D50~\cite{feichtenhofer2019slowfast} and video transformer ViT\cite{dosovitskiy2020vit}}, as shown in Table~\ref{tab:backbone}. In these experiments, we simply replicate the condensed frames multiple times along the the temporal dimension to meet the 3D models and Transformers. From the results, we can find that our FrameMaker can consistently achieve comparable performance on the three different backbones with only 12\% memory cost. These results demonstrate the better generalization of FrameMaker.

\paragraph{Alternative prompting strategies.} We conduct experiments to explore the types and positions of the prompts in Table~\ref{tab:prompt_strategies}. \textit{(i)}
We compare our Instance-Specific (I.-Spec.) Prompt (ISP) with the Task- (T.-Spec.) and Class-Specific (C.-Spec.) prompts and observe that our ISP can achieve the best performance when operating on both features and frames. 
Since the actions in videos have great intra-class variance, simply sharing task and class-specific prompts is insufficient to reserve the important spatia-temporal cues for each instance; \textit{(ii) Positions of prompt}: 
For the feature settings, we add the prompt on the features of the 4 stages of the ResNet. The feature-based prompt is worse than our frame-based prompt, which may be caused by the mismatch between the running model and fixed prompts during incremental procedure. 
Moreover, additional features require additional storage space because the corresponding prompt cannot be directly added on the changed feature.

\begin{table}
  \caption{Analysis for memory budget on UCF101 with 10 steps. Here `$x\text{F}\times y\text{V}$=$z$Mb' indicates that $x$ sampled frames from $y$ different videos are stored for each class, and $x$ frames are sampled from each video. We assume that the spatial resolution of frames is 224$\times$224, and the total memory consumption is $z$Mbytes.}
  \label{tab:memory_budget}
  \centering
  \begin{tabular}{ccccccc}
    \toprule[1.15pt]
    Memory Per Class & \multicolumn{2}{c}{$8\text{F}\times1\text{V}$=1.2Mb} & \multicolumn{2}{c}{$8\text{F}\times2\text{V}$=2.4Mb} & \multicolumn{2}{c}{$8\text{F}\times5\text{V}$=6.0Mb}  \\
    Classifier  & CNN & NME & CNN & NME & CNN & NME\\
    \cmidrule(r){1-1}\cmidrule(r){2-3}\cmidrule(r){4-5}\cmidrule(r){6-7}
    iCaRL~\cite{rebuffi2017icarl} & - & 58.05 & - & 60.50 & - & 64.51 \\
    UCIR~\cite{hou2019ucir} & 61.92 & 65.52 & 66.43 & 67.58 & 70.42 & 70.50  \\
    PODNet~\cite{douillard2020podnet} & 63.18 & 70.96 & 65.93 & 72.78 & 71.58 & 73.75   \\
    TCD~\cite{park2021tcd} & 64.52 & 71.96 & 68.40 & 73.30 & 73.43 & 75.35   \\
    \cmidrule(r){1-1}\cmidrule(r){2-3}\cmidrule(r){4-5}\cmidrule(r){6-7}
    Memory Per Class & \multicolumn{2}{c}{$1\text{F}\times1\text{V}$=0.15Mb} & \multicolumn{2}{c}{$1\text{F}\times2\text{V}$=0.3Mb}& \multicolumn{2}{c}{$1\text{F}\times5\text{V}$=0.75Mb}  \\
    \cmidrule(r){1-1}\cmidrule(r){2-3}\cmidrule(r){4-5}\cmidrule(r){6-7}
    FrameMaker & 49.37 & 70.78 & 62.06 & 74.18 & 72.93 & 76.64  \\
    \cmidrule(r){1-1}\cmidrule(r){2-3}\cmidrule(r){4-5}\cmidrule(r){6-7}
    Memory Per Class & \multicolumn{2}{c}{$1\text{F}\times8\text{V}$=1.2Mb} & \multicolumn{2}{c}{$1\text{F}\times16\text{V}$=2.4Mb}& \multicolumn{2}{c}{$1\text{F}\times40\text{V}$=6.0Mb}  \\
    \cmidrule(r){1-1}\cmidrule(r){2-3}\cmidrule(r){4-5}\cmidrule(r){6-7}
    FrameMaker & \textbf{73.64} & \textbf{76.98} & \textbf{75.19} & \textbf{77.43} & \textbf{76.38} & \textbf{78.14} \\
    \bottomrule[1.15pt]
  \end{tabular}
\end{table}

\paragraph{Memory budget comparison.}
We now compare the memory budget of FramMaker with existing approaches. 
To make a direct comparison, we use the definition of the working memory size in terms of stored frames, following~\cite{villa2022vclimb}.
The results are summarized in Table~\ref{tab:memory_budget}. We can see that our FrameMaker only costs 1.2Mb memory, which surpasses TCD~\cite{park2021tcd} using 6.0Mb by 0.21\% and 1.63\% in terms of CNN and NME. 
FrameMaker saves 80\% of memory space with the comparable performance.
Meanwhile, we further increase the stored videos but keep the same memory with existing methods.
FrameMaker can still effectively promote performance and shows superior abilities to fight catastrophic forgetting.
Our results are about 3\% ahead of TCD in both indicators.
Besides, we also attempt to select the same number of videos as the existing methods.
However, the accuracy of the trained linear classifier, \textit{i.e.,} CNN, is weaker with a small number of video instances. 
Interestingly, NME evaluated by the exemplar class centre yielded fairly or even better results, implying that the model suffers only slight forgetting.
We hypothesize that this is caused by the poor feature diversity that a few condensed frames are provided.
Therefore, the classifier tends to overfit fixed sample features, resulting in poor generalization.

\paragraph{Visualization of condensing weights and instance-specific prompts.}

To intuitively understand our FrameMaker, we provide visualization of condensing weights, Instance-Specific Prompts, condensed frames and learned prompts in Figure~\ref{fig:vis_weight_prompt}. 
To save space, we uniformly select the most typical 4 frames from $T$ frames for display.
Based on this, we have the following fascinating observations:
(1) For frames with strong scene bias, the learned weights are more uniform, while those with motion bias are not. 
For example, the action classification for (a) and (b) in Figure~\ref{fig:vis_weight_prompt} does not depend on a specific frame. Hence, the condensing weight for each frame is roughly the same.
In contrast, \textit{stair climbing} and \textit{hugging} in (c) and (d) only occur in some specific frames, and those keyframes are highlighted by condensing weights.
This demonstrates that it is reasonable to condense multiple frames into one frame since our proposed Frame Condensing can assist the learning of which frame is meaningful for the action recognition task.
(2) It is difficult for humans to absorb useful knowledge from the learned prompts shown in Figure~\ref{fig:vis_weight_prompt}, and even they seem to be the same. 
It should be emphasized that the prompts are different. 
We try to share the same prompt for all videos to verify this.
The experiment is conducted on UCF101 with 10 steps, and the performances decrease by 0.83\% and 0.78\% in terms of CNN and NME, respectively.
This fully validates that our Instance-Specific Prompt is to replenish the specific information for different instances.


\begin{figure}
  \centering
  \includegraphics[width=1.0\linewidth]{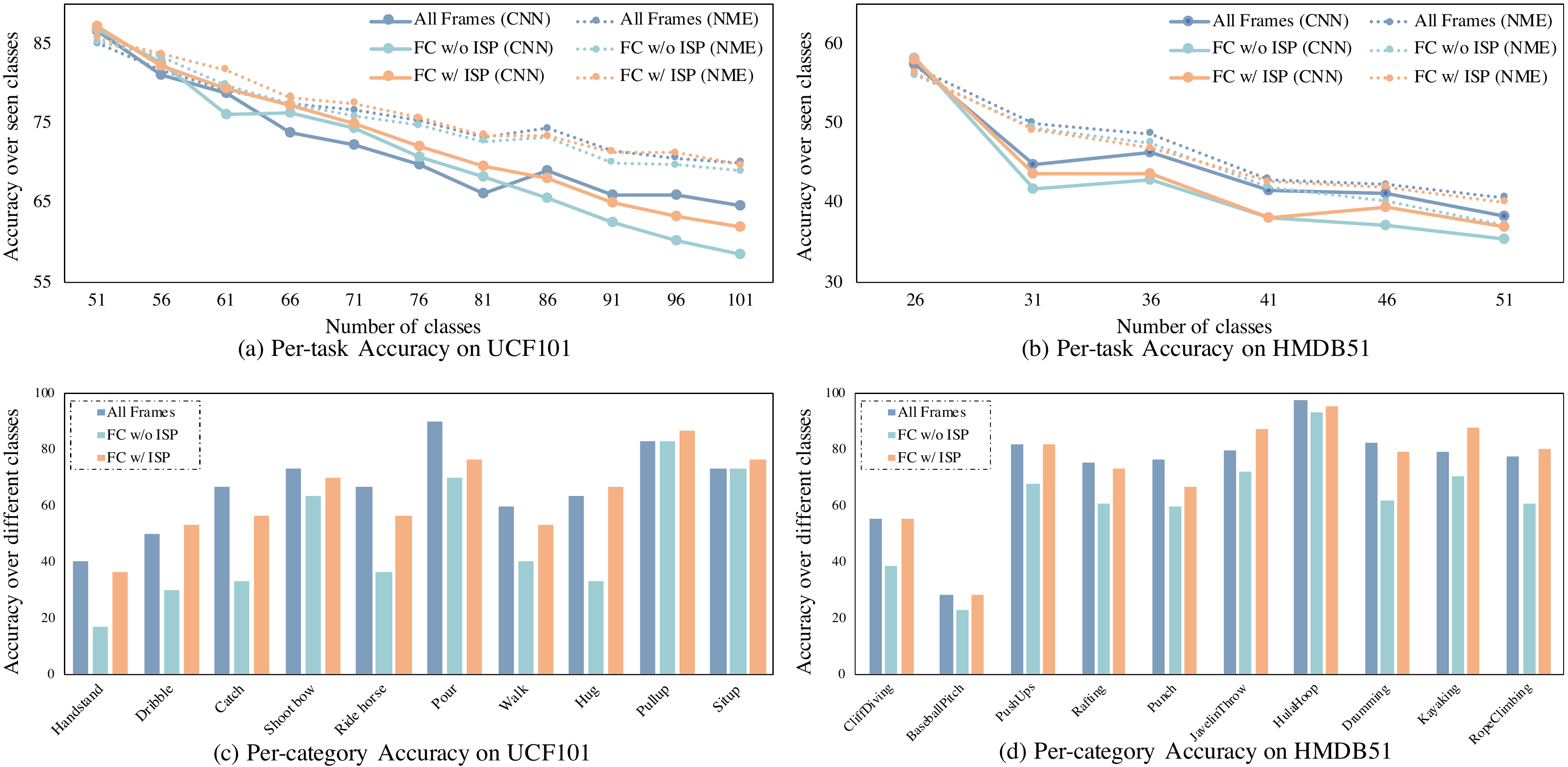}
  \caption{Per-category and per-task accuracy on UCF101 and HMDB51. `FC' and `ISP' refer to Frame Condensing and Instance-Specific Prompt, respectively. The plots in (a) and (b) indicate that the introduction of prompt improves accuracy in almost all tasks. The results in (c) and (d) show that our Instance-Specific Prompt significantly improves performance from Frame Condensing in all categories on both datasets.}
  \label{fig:class_accuracy}
\end{figure}
\paragraph{Per-category and per-task analysis.} We depict the task-wise and category-wise accuracy with different forms of stored frames in Figure~\ref{fig:class_accuracy}, \ie, "\textit{All Frames}", "\textit{Frame Condensing without Instance-Specific Prompt}" and "\textit{Frame Condensing with Instance-Specific Prompt}". 
As expected, integrating multiple frames into one condensed frame does undermine performance in almost all cases unless Instance-Specific Prompt is introduced.
%
As shown in Figure~\ref{fig:class_accuracy}(a) and (b), the gains on incremental tasks yielded by Instance-Specific Prompt mainly fall in the later tasks, which implies that the spatio-temporal knowledge absorbed by prompt can effectively alleviate forgetting.
Meanwhile, from Figure~\ref{fig:class_accuracy}(c) and (d), we also observe that Frame Condensing can lead to the performance degradation of actions with large temporal and spatial variations or short duration, such as "\textit{CliffDiving}", "\textit{Ride horse}" and "\textit{Hug}", etc.
Therefore, Frame Condensing with only a few parameters is difficult to capture these complex spatio-temporal interactions.
However, Instance-Specific Prompt helps alleviate this issue since it recovers the lost spatio-temporal details in Frame Condensing for each instance.

\section{Discussions}

\paragraph{Limitations.} Compared with the previous video incremental learning approaches, FrameMaker provides a simple framework that significantly reduces the amount of memory required for each exemplar video. Nevertheless, FrameMaker may fail when the memory budget is highly constrained, which needs to be further explored in future work.

\paragraph{Societal Impacts.} Although FrameMaker only stores condensed frames in memory buffer for experience replay, which still retains some information about the original video. Therefore, FrameMaker may be used in applications with privacy concerns~\cite{shokri2015privacy}.

\paragraph{Conclusion.} 
This paper proposes FrameMaker, a memory-efficient approach for video class-incremental learning. 
It explores how to learn an effective condensed frame with less memory for each video that can be applied to the future incremental task. 
FrameMaker mainly consists of two key components, \textit{i.e.,} \textit{Frame Condensing} and \textit{Instance-Specific Prompt}. 
%
%
The former learns a group of condensed weights for each video to integrate multiple frames into one condensed frame, while the latter learns to retrieve the collapsed spatio-temporal structure for the condensed frame.
%
%
In this way, FrameMaker offers better results with only 20\% memory consumption compared to recent advanced methods and sets a new state-of-the-art class-incremental learning performance on multiple challenging datasets. 
%
%
%
%
Memory-based class-incremental learning is one of the effective methods to fight forgetting. 
The efficient utilization of limited space is worth exploring in future work.

\paragraph{Acknowledgments.}
This work is supported by  the National Key R\&D Program of China under Grant No. 2018AAA0101501, the National Natural Science Foundation of China undergrant 61871435 and 62272380, Fundamental Research Funds for the Central Universities no.2019kfyXKJC024, the Science and Technology Program of Xi’an, China under Grant 21RGZN0017, and by Alibaba Group through Alibaba Innovative Research Program.

\bibliographystyle{plainnat}
\bibliography{egbib}

\begin{thebibliography}{67}
\providecommand{\natexlab}[1]{#1}
\providecommand{\url}[1]{\texttt{#1}}
\expandafter\ifx\csname urlstyle\endcsname\relax
  \providecommand{\doi}[1]{doi: #1}\else
  \providecommand{\doi}{doi: \begingroup \urlstyle{rm}\Url}\fi

\bibitem[Aljundi et~al.(2018)Aljundi, Babiloni, Elhoseiny, Rohrbach, and
  Tuytelaars]{aljundi2018mas}
Rahaf Aljundi, Francesca Babiloni, Mohamed Elhoseiny, Marcus Rohrbach, and
  Tinne Tuytelaars.
\newblock Memory aware synapses: Learning what (not) to forget.
\newblock In \emph{Proceedings of the European Conference on Computer Vision},
  pages 139--154, 2018.

\bibitem[Arnab et~al.(2021)Arnab, Dehghani, Heigold, Sun, Lu{\v{c}}i{\'c}, and
  Schmid]{arnab2021vivit}
Anurag Arnab, Mostafa Dehghani, Georg Heigold, Chen Sun, Mario Lu{\v{c}}i{\'c},
  and Cordelia Schmid.
\newblock Vivit: A video vision transformer.
\newblock In \emph{Proceedings of the IEEE/CVF International Conference on
  Computer Vision}, pages 6836--6846, 2021.

\bibitem[Bahng et~al.(2022)Bahng, Jahanian, Sankaranarayanan, and
  Isola]{bahng2022visual_prompt_pixel}
Hyojin Bahng, Ali Jahanian, Swami Sankaranarayanan, and Phillip Isola.
\newblock Visual prompting: Modifying pixel space to adapt pre-trained models.
\newblock \emph{arXiv preprint arXiv:2203.17274}, 2022.

\bibitem[Bertasius et~al.(2021)Bertasius, Wang, and
  Torresani]{bertasius2021timesformer}
Gedas Bertasius, Heng Wang, and Lorenzo Torresani.
\newblock Is space-time attention all you need for video understanding.
\newblock \emph{arXiv preprint arXiv:2102.05095}, 2\penalty0 (3):\penalty0 4,
  2021.

\bibitem[Buzzega et~al.(2020)Buzzega, Boschini, Porrello, Abati, and
  Calderara]{buzzega2020glc}
Pietro Buzzega, Matteo Boschini, Angelo Porrello, Davide Abati, and Simone
  Calderara.
\newblock Dark experience for general continual learning: a strong, simple
  baseline.
\newblock \emph{Advances in neural information processing systems},
  33:\penalty0 15920--15930, 2020.

\bibitem[Carreira and Zisserman(2017)]{carreira2017i3d}
Joao Carreira and Andrew Zisserman.
\newblock Quo vadis, action recognition? a new model and the kinetics dataset.
\newblock In \emph{proceedings of the IEEE Conference on Computer Vision and
  Pattern Recognition}, pages 6299--6308, 2017.

\bibitem[Castro et~al.(2018)Castro, Mar{\'\i}n-Jim{\'e}nez, Guil, Schmid, and
  Alahari]{castro2018e2eil}
Francisco~M Castro, Manuel~J Mar{\'\i}n-Jim{\'e}nez, Nicol{\'a}s Guil, Cordelia
  Schmid, and Karteek Alahari.
\newblock End-to-end incremental learning.
\newblock In \emph{Proceedings of the European conference on computer vision},
  pages 233--248, 2018.

\bibitem[Deng et~al.(2009)Deng, Dong, Socher, Li, Li, and
  Fei-Fei]{deng2009imagenet}
Jia Deng, Wei Dong, Richard Socher, Li-Jia Li, Kai Li, and Li~Fei-Fei.
\newblock Imagenet: A large-scale hierarchical image database.
\newblock In \emph{2009 IEEE conference on computer vision and pattern
  recognition}, pages 248--255. Ieee, 2009.

\bibitem[Dhar et~al.(2019)Dhar, Singh, Peng, Wu, and Chellappa]{dhar2019lwm}
Prithviraj Dhar, Rajat~Vikram Singh, Kuan-Chuan Peng, Ziyan Wu, and Rama
  Chellappa.
\newblock Learning without memorizing.
\newblock In \emph{Proceedings of the IEEE/CVF Conference on Computer Vision
  and Pattern Recognition}, pages 5138--5146, 2019.

\bibitem[Diba et~al.(2018)Diba, Fayyaz, Sharma, Arzani, Yousefzadeh, Gall, and
  Van~Gool]{diba2018stc}
Ali Diba, Mohsen Fayyaz, Vivek Sharma, M~Mahdi Arzani, Rahman Yousefzadeh,
  Juergen Gall, and Luc Van~Gool.
\newblock Spatio-temporal channel correlation networks for action
  classification.
\newblock In \emph{Proceedings of the European Conference on Computer Vision},
  pages 284--299, 2018.

\bibitem[Dosovitskiy et~al.(2020)Dosovitskiy, Beyer, Kolesnikov, Weissenborn,
  Zhai, Unterthiner, Dehghani, Minderer, Heigold, Gelly,
  et~al.]{dosovitskiy2020vit}
Alexey Dosovitskiy, Lucas Beyer, Alexander Kolesnikov, Dirk Weissenborn,
  Xiaohua Zhai, Thomas Unterthiner, Mostafa Dehghani, Matthias Minderer, Georg
  Heigold, Sylvain Gelly, et~al.
\newblock An image is worth 16x16 words: Transformers for image recognition at
  scale.
\newblock \emph{arXiv preprint arXiv:2010.11929}, 2020.

\bibitem[Douillard et~al.(2020)Douillard, Cord, Ollion, Robert, and
  Valle]{douillard2020podnet}
Arthur Douillard, Matthieu Cord, Charles Ollion, Thomas Robert, and Eduardo
  Valle.
\newblock Podnet: Pooled outputs distillation for small-tasks incremental
  learning.
\newblock In \emph{European Conference on Computer Vision}, pages 86--102.
  Springer, 2020.

\bibitem[Fan et~al.(2021)Fan, Xiong, Mangalam, Li, Yan, Malik, and
  Feichtenhofer]{fan2021mvit}
Haoqi Fan, Bo~Xiong, Karttikeya Mangalam, Yanghao Li, Zhicheng Yan, Jitendra
  Malik, and Christoph Feichtenhofer.
\newblock Multiscale vision transformers.
\newblock In \emph{Proceedings of the IEEE/CVF International Conference on
  Computer Vision}, pages 6824--6835, 2021.

\bibitem[Fan et~al.(2018)Fan, Xu, Zhu, Yan, Ge, and Yang]{fan2018watching}
Hehe Fan, Zhongwen Xu, Linchao Zhu, Chenggang Yan, Jianjun Ge, and Yi~Yang.
\newblock Watching a small portion could be as good as watching all: Towards
  efficient video classification.
\newblock In \emph{IJCAI International Joint Conference on Artificial
  Intelligence}, 2018.

\bibitem[Feichtenhofer(2020)]{feichtenhofer2020x3d}
Christoph Feichtenhofer.
\newblock X3d: Expanding architectures for efficient video recognition.
\newblock In \emph{Proceedings of the IEEE/CVF Conference on Computer Vision
  and Pattern Recognition}, pages 203--213, 2020.

\bibitem[Feichtenhofer et~al.(2019)Feichtenhofer, Fan, Malik, and
  He]{feichtenhofer2019slowfast}
Christoph Feichtenhofer, Haoqi Fan, Jitendra Malik, and Kaiming He.
\newblock Slowfast networks for video recognition.
\newblock In \emph{Proceedings of the IEEE/CVF international conference on
  computer vision}, pages 6202--6211, 2019.

\bibitem[Gowda et~al.(2021)Gowda, Rohrbach, and Sevilla-Lara]{gowda2021smart}
Shreyank~N Gowda, Marcus Rohrbach, and Laura Sevilla-Lara.
\newblock Smart frame selection for action recognition.
\newblock In \emph{Proceedings of the AAAI Conference on Artificial
  Intelligence}, volume~35, pages 1451--1459, 2021.

\bibitem[Goyal et~al.(2017)Goyal, Ebrahimi~Kahou, Michalski, Materzynska,
  Westphal, Kim, Haenel, Fruend, Yianilos, Mueller-Freitag,
  et~al.]{goyal2017something}
Raghav Goyal, Samira Ebrahimi~Kahou, Vincent Michalski, Joanna Materzynska,
  Susanne Westphal, Heuna Kim, Valentin Haenel, Ingo Fruend, Peter Yianilos,
  Moritz Mueller-Freitag, et~al.
\newblock The" something something" video database for learning and evaluating
  visual common sense.
\newblock In \emph{Proceedings of the IEEE international conference on computer
  vision}, pages 5842--5850, 2017.

\bibitem[He et~al.(2016)He, Zhang, Ren, and Sun]{he2016resnet}
Kaiming He, Xiangyu Zhang, Shaoqing Ren, and Jian Sun.
\newblock Deep residual learning for image recognition.
\newblock In \emph{Proceedings of the IEEE conference on computer vision and
  pattern recognition}, pages 770--778, 2016.

\bibitem[Hinton et~al.(2015)Hinton, Vinyals, Dean,
  et~al.]{hinton2015distilling}
Geoffrey Hinton, Oriol Vinyals, Jeff Dean, et~al.
\newblock Distilling the knowledge in a neural network.
\newblock \emph{arXiv preprint arXiv:1503.02531}, 2\penalty0 (7), 2015.

\bibitem[Hou et~al.(2019)Hou, Pan, Loy, Wang, and Lin]{hou2019ucir}
Saihui Hou, Xinyu Pan, Chen~Change Loy, Zilei Wang, and Dahua Lin.
\newblock Learning a unified classifier incrementally via rebalancing.
\newblock In \emph{Proceedings of the IEEE/CVF Conference on Computer Vision
  and Pattern Recognition}, pages 831--839, 2019.

\bibitem[Hu et~al.(2019)Hu, Zhang, Xie, and Lin]{hu2019local}
Han Hu, Zheng Zhang, Zhenda Xie, and Stephen Lin.
\newblock Local relation networks for image recognition.
\newblock In \emph{Proceedings of the IEEE/CVF International Conference on
  Computer Vision}, 2019.

\bibitem[Huang et~al.(2018)Huang, Ramanathan, Mahajan, Torresani, Paluri,
  Fei-Fei, and Niebles]{huang2018makes}
De-An Huang, Vignesh Ramanathan, Dhruv Mahajan, Lorenzo Torresani, Manohar
  Paluri, Li~Fei-Fei, and Juan~Carlos Niebles.
\newblock What makes a video a video: Analyzing temporal information in video
  understanding models and datasets.
\newblock In \emph{Proceedings of the IEEE Conference on Computer Vision and
  Pattern Recognition}, pages 7366--7375, 2018.

\bibitem[Jia et~al.(2022)Jia, Tang, Chen, Cardie, Belongie, Hariharan, and
  Lim]{jia2022visual_prompt_token}
Menglin Jia, Luming Tang, Bor-Chun Chen, Claire Cardie, Serge Belongie, Bharath
  Hariharan, and Ser-Nam Lim.
\newblock Visual prompt tuning.
\newblock \emph{arXiv preprint arXiv:2203.12119}, 2022.

\bibitem[Kim et~al.(2022)Kim, Gowda, Mac~Aodha, and
  Sevilla-Lara]{kim2022greyst}
Kiyoon Kim, Shreyank~N Gowda, Oisin Mac~Aodha, and Laura Sevilla-Lara.
\newblock Capturing temporal information in a single frame: Channel sampling
  strategies for action recognition.
\newblock \emph{arXiv preprint arXiv:2201.10394}, 2022.

\bibitem[Kirkpatrick et~al.(2017)Kirkpatrick, Pascanu, Rabinowitz, Veness,
  Desjardins, Rusu, Milan, Quan, Ramalho, Grabska-Barwinska,
  et~al.]{kirkpatrick2017img_reg2}
James Kirkpatrick, Razvan Pascanu, Neil Rabinowitz, Joel Veness, Guillaume
  Desjardins, Andrei~A Rusu, Kieran Milan, John Quan, Tiago Ramalho, Agnieszka
  Grabska-Barwinska, et~al.
\newblock Overcoming catastrophic forgetting in neural networks.
\newblock \emph{Proceedings of the national academy of sciences}, 114\penalty0
  (13):\penalty0 3521--3526, 2017.

\bibitem[Korbar et~al.(2019)Korbar, Tran, and Torresani]{korbar2019scsampler}
Bruno Korbar, Du~Tran, and Lorenzo Torresani.
\newblock Scsampler: Sampling salient clips from video for efficient action
  recognition.
\newblock In \emph{Proceedings of the IEEE/CVF International Conference on
  Computer Vision}, pages 6232--6242, 2019.

\bibitem[Kuehne et~al.(2011)Kuehne, Jhuang, Garrote, Poggio, and
  Serre]{kuehne2011hmdb}
Hildegard Kuehne, Hueihan Jhuang, Est{\'\i}baliz Garrote, Tomaso Poggio, and
  Thomas Serre.
\newblock Hmdb: a large video database for human motion recognition.
\newblock In \emph{2011 International conference on computer vision}, pages
  2556--2563. IEEE, 2011.

\bibitem[Lester et~al.(2021)Lester, Al-Rfou, and
  Constant]{lester2021learnable_prompt1}
Brian Lester, Rami Al-Rfou, and Noah Constant.
\newblock The power of scale for parameter-efficient prompt tuning.
\newblock \emph{arXiv preprint arXiv:2104.08691}, 2021.

\bibitem[Li et~al.(2022)Li, Wang, Gao, Song, Liu, Li, and
  Qiao]{li2022uniformer}
Kunchang Li, Yali Wang, Peng Gao, Guanglu Song, Yu~Liu, Hongsheng Li, and
  Yu~Qiao.
\newblock Uniformer: Unified transformer for efficient spatiotemporal
  representation learning.
\newblock \emph{arXiv preprint arXiv:2201.04676}, 2022.

\bibitem[Li and Liang(2021)]{li2021learnable_prompt2}
Xiang~Lisa Li and Percy Liang.
\newblock Prefix-tuning: Optimizing continuous prompts for generation.
\newblock \emph{arXiv preprint arXiv:2101.00190}, 2021.

\bibitem[Li and Hoiem(2017{\natexlab{a}})]{li2017img_reg4}
Zhizhong Li and Derek Hoiem.
\newblock Learning without forgetting.
\newblock \emph{IEEE transactions on pattern analysis and machine
  intelligence}, 40\penalty0 (12):\penalty0 2935--2947, 2017{\natexlab{a}}.

\bibitem[Li and Hoiem(2017{\natexlab{b}})]{li2017lwfmc}
Zhizhong Li and Derek Hoiem.
\newblock Learning without forgetting.
\newblock \emph{IEEE transactions on pattern analysis and machine
  intelligence}, 40\penalty0 (12):\penalty0 2935--2947, 2017{\natexlab{b}}.

\bibitem[Lin et~al.(2019)Lin, Gan, and Han]{lin2019tsm}
Ji~Lin, Chuang Gan, and Song Han.
\newblock Tsm: Temporal shift module for efficient video understanding.
\newblock In \emph{Proceedings of the IEEE/CVF International Conference on
  Computer Vision}, pages 7083--7093, 2019.

\bibitem[Liu et~al.(2022{\natexlab{a}})Liu, Yang, Fu, and
  Qian]{liu2022learning}
Chengxu Liu, Huan Yang, Jianlong Fu, and Xueming Qian.
\newblock Learning trajectory-aware transformer for video super-resolution.
\newblock In \emph{Proceedings of the IEEE/CVF Conference on Computer Vision
  and Pattern Recognition}, 2022{\natexlab{a}}.

\bibitem[Liu et~al.(2021{\natexlab{a}})Liu, Yuan, Fu, Jiang, Hayashi, and
  Neubig]{liu2021prompt}
Pengfei Liu, Weizhe Yuan, Jinlan Fu, Zhengbao Jiang, Hiroaki Hayashi, and
  Graham Neubig.
\newblock Pre-train, prompt, and predict: A systematic survey of prompting
  methods in natural language processing.
\newblock \emph{arXiv preprint arXiv:2107.13586}, 2021{\natexlab{a}}.

\bibitem[Liu et~al.(2021{\natexlab{b}})Liu, Ning, Cao, Wei, Zhang, Lin, and
  Hu]{liu2021videoswin}
Ze~Liu, Jia Ning, Yue Cao, Yixuan Wei, Zheng Zhang, Stephen Lin, and Han Hu.
\newblock Video swin transformer.
\newblock \emph{arXiv preprint arXiv:2106.13230}, 2021{\natexlab{b}}.

\bibitem[Liu et~al.(2022{\natexlab{b}})Liu, Hu, Lin, Yao, Xie, Wei, Ning, Cao,
  Zhang, Dong, Wei, and Guo]{liu2021swinv2}
Ze~Liu, Han Hu, Yutong Lin, Zhuliang Yao, Zhenda Xie, Yixuan Wei, Jia Ning, Yue
  Cao, Zheng Zhang, Li~Dong, Furu Wei, and Baining Guo.
\newblock Swin transformer v2: Scaling up capacity and resolution.
\newblock In \emph{International Conference on Computer Vision and Pattern
  Recognition}, 2022{\natexlab{b}}.

\bibitem[McCloskey and Cohen(1989)]{mccloskey1989catastrophic}
Michael McCloskey and Neal~J Cohen.
\newblock Catastrophic interference in connectionist networks: The sequential
  learning problem.
\newblock In \emph{Psychology of learning and motivation}, volume~24, pages
  109--165. Elsevier, 1989.

\bibitem[Ostapenko et~al.(2019)Ostapenko, Puscas, Klein, Jahnichen, and
  Nabi]{ostapenko2019gen1}
Oleksiy Ostapenko, Mihai Puscas, Tassilo Klein, Patrick Jahnichen, and Moin
  Nabi.
\newblock Learning to remember: A synaptic plasticity driven framework for
  continual learning.
\newblock In \emph{Proceedings of the IEEE/CVF Conference on Computer Vision
  and Pattern Recognition}, pages 11321--11329, 2019.

\bibitem[Park et~al.(2021)Park, Kang, and Han]{park2021tcd}
Jaeyoo Park, Minsoo Kang, and Bohyung Han.
\newblock Class-incremental learning for action recognition in videos.
\newblock In \emph{Proceedings of the IEEE/CVF International Conference on
  Computer Vision}, pages 13698--13707, 2021.

\bibitem[Paszke et~al.(2019)Paszke, Gross, Massa, Lerer, Bradbury, Chanan,
  Killeen, Lin, Gimelshein, Antiga, et~al.]{paszke2019pytorch}
Adam Paszke, Sam Gross, Francisco Massa, Adam Lerer, James Bradbury, Gregory
  Chanan, Trevor Killeen, Zeming Lin, Natalia Gimelshein, Luca Antiga, et~al.
\newblock Pytorch: An imperative style, high-performance deep learning library.
\newblock \emph{Advances in neural information processing systems}, 32, 2019.

\bibitem[Qiu et~al.(2017)Qiu, Yao, and Mei]{qiu2017p3d}
Zhaofan Qiu, Ting Yao, and Tao Mei.
\newblock Learning spatio-temporal representation with pseudo-3d residual
  networks.
\newblock In \emph{proceedings of the IEEE International Conference on Computer
  Vision}, pages 5533--5541, 2017.

\bibitem[Qiu et~al.(2021)Qiu, Yao, Shu, Ngo, and Mei]{qiu2021condensing}
Zhaofan Qiu, Ting Yao, Yan Shu, Chong-Wah Ngo, and Tao Mei.
\newblock Condensing a sequence to one informative frame for video recognition.
\newblock In \emph{Proceedings of the IEEE/CVF International Conference on
  Computer Vision}, pages 16311--16320, 2021.

\bibitem[Rebuffi et~al.(2017)Rebuffi, Kolesnikov, Sperl, and
  Lampert]{rebuffi2017icarl}
Sylvestre-Alvise Rebuffi, Alexander Kolesnikov, Georg Sperl, and Christoph~H
  Lampert.
\newblock icarl: Incremental classifier and representation learning.
\newblock In \emph{Proceedings of the IEEE conference on Computer Vision and
  Pattern Recognition}, pages 2001--2010, 2017.

\bibitem[Reddy and Shah(2013)]{reddy2013recognizing}
Kishore~K Reddy and Mubarak Shah.
\newblock Recognizing 50 human action categories of web videos.
\newblock \emph{Machine vision and applications}, 24\penalty0 (5):\penalty0
  971--981, 2013.

\bibitem[Shin et~al.(2017)Shin, Lee, Kim, and Kim]{shin2017gen2}
Hanul Shin, Jung~Kwon Lee, Jaehong Kim, and Jiwon Kim.
\newblock Continual learning with deep generative replay.
\newblock \emph{Advances in neural information processing systems}, 30, 2017.

\bibitem[Shokri and Shmatikov(2015)]{shokri2015privacy}
Reza Shokri and Vitaly Shmatikov.
\newblock Privacy-preserving deep learning.
\newblock In \emph{Proceedings of the 22nd ACM SIGSAC conference on computer
  and communications security}, pages 1310--1321, 2015.

\bibitem[Simonyan and Zisserman(2014)]{simonyan2014vgg}
Karen Simonyan and Andrew Zisserman.
\newblock Very deep convolutional networks for large-scale image recognition.
\newblock \emph{arXiv preprint arXiv:1409.1556}, 2014.

\bibitem[Soomro et~al.(2012)Soomro, Zamir, and Shah]{soomro2012ucf101}
Khurram Soomro, Amir~Roshan Zamir, and Mubarak Shah.
\newblock Ucf101: A dataset of 101 human actions classes from videos in the
  wild.
\newblock \emph{arXiv preprint arXiv:1212.0402}, 2012.

\bibitem[Tran et~al.(2015)Tran, Bourdev, Fergus, Torresani, and
  Paluri]{tran2015c3d}
Du~Tran, Lubomir Bourdev, Rob Fergus, Lorenzo Torresani, and Manohar Paluri.
\newblock Learning spatiotemporal features with 3d convolutional networks.
\newblock In \emph{Proceedings of the IEEE international conference on computer
  vision}, pages 4489--4497, 2015.

\bibitem[Tran et~al.(2018)Tran, Wang, Torresani, Ray, LeCun, and
  Paluri]{tran2018r21d}
Du~Tran, Heng Wang, Lorenzo Torresani, Jamie Ray, Yann LeCun, and Manohar
  Paluri.
\newblock A closer look at spatiotemporal convolutions for action recognition.
\newblock In \emph{Proceedings of the IEEE conference on Computer Vision and
  Pattern Recognition}, pages 6450--6459, 2018.

\bibitem[Tran et~al.(2019)Tran, Wang, Torresani, and Feiszli]{tran2019csn}
Du~Tran, Heng Wang, Lorenzo Torresani, and Matt Feiszli.
\newblock Video classification with channel-separated convolutional networks.
\newblock In \emph{Proceedings of the IEEE/CVF International Conference on
  Computer Vision}, pages 5552--5561, 2019.

\bibitem[Villa et~al.(2022)Villa, Alhamoud, Alc{\'a}zar, Heilbron, Escorcia,
  and Ghanem]{villa2022vclimb}
Andr{\'e}s Villa, Kumail Alhamoud, Juan~Le{\'o}n Alc{\'a}zar, Fabian~Caba
  Heilbron, Victor Escorcia, and Bernard Ghanem.
\newblock vclimb: A novel video class incremental learning benchmark.
\newblock \emph{arXiv preprint arXiv:2201.09381}, 2022.

\bibitem[Wang et~al.(2019)Wang, Wang, Wang, Wang, Feng, and
  Tan]{wang2019stacked}
Junbo Wang, Wei Wang, Zhiyong Wang, Liang Wang, Dagan Feng, and Tieniu Tan.
\newblock Stacked memory network for video summarization.
\newblock In \emph{Proceedings of the 27th ACM International Conference on
  Multimedia}, pages 836--844, 2019.

\bibitem[Wang et~al.(2018)Wang, Li, Li, and Van~Gool]{wang2018atrnet}
Limin Wang, Wei Li, Wen Li, and Luc Van~Gool.
\newblock Appearance-and-relation networks for video classification.
\newblock In \emph{Proceedings of the IEEE conference on computer vision and
  pattern recognition}, pages 1430--1439, 2018.

\bibitem[Wang et~al.(2021)Wang, Zhang, Lee, Zhang, Sun, Ren, Su, Perot, Dy, and
  Pfister]{wang2021l2p}
Zifeng Wang, Zizhao Zhang, Chen-Yu Lee, Han Zhang, Ruoxi Sun, Xiaoqi Ren,
  Guolong Su, Vincent Perot, Jennifer Dy, and Tomas Pfister.
\newblock Learning to prompt for continual learning.
\newblock \emph{arXiv preprint arXiv:2112.08654}, 2021.

\bibitem[Wang et~al.(2022)Wang, Zhang, Ebrahimi, Sun, Zhang, Lee, Ren, Su,
  Perot, Dy, et~al.]{wang2022dualprompt}
Zifeng Wang, Zizhao Zhang, Sayna Ebrahimi, Ruoxi Sun, Han Zhang, Chen-Yu Lee,
  Xiaoqi Ren, Guolong Su, Vincent Perot, Jennifer Dy, et~al.
\newblock Dualprompt: Complementary prompting for rehearsal-free continual
  learning.
\newblock \emph{arXiv preprint arXiv:2204.04799}, 2022.

\bibitem[Wu et~al.(2019{\natexlab{a}})Wu, He, Tan, Chen, and Wen]{wu2019multi}
Wenhao Wu, Dongliang He, Xiao Tan, Shifeng Chen, and Shilei Wen.
\newblock Multi-agent reinforcement learning based frame sampling for effective
  untrimmed video recognition.
\newblock In \emph{Proceedings of the IEEE/CVF International Conference on
  Computer Vision}, pages 6222--6231, 2019{\natexlab{a}}.

\bibitem[Wu et~al.(2019{\natexlab{b}})Wu, Chen, Wang, Ye, Liu, Guo, and
  Fu]{wu2019bic}
Yue Wu, Yinpeng Chen, Lijuan Wang, Yuancheng Ye, Zicheng Liu, Yandong Guo, and
  Yun Fu.
\newblock Large scale incremental learning.
\newblock In \emph{Proceedings of the IEEE/CVF Conference on Computer Vision
  and Pattern Recognition}, pages 374--382, 2019{\natexlab{b}}.

\bibitem[Wu et~al.(2019{\natexlab{c}})Wu, Xiong, Ma, Socher, and
  Davis]{wu2019adaframe}
Zuxuan Wu, Caiming Xiong, Chih-Yao Ma, Richard Socher, and Larry~S Davis.
\newblock Adaframe: Adaptive frame selection for fast video recognition.
\newblock In \emph{Proceedings of the IEEE/CVF Conference on Computer Vision
  and Pattern Recognition}, pages 1278--1287, 2019{\natexlab{c}}.

\bibitem[Xie et~al.(2017)Xie, Sun, Huang, Tu, and Murphy]{xie2017s3d}
Saining Xie, Chen Sun, Jonathan Huang, Zhuowen Tu, and Kevin Murphy.
\newblock Rethinking spatiotemporal feature learning for video understanding.
\newblock \emph{arXiv preprint arXiv:1712.04851}, 1\penalty0 (2):\penalty0 5,
  2017.

\bibitem[Zenke et~al.(2017)Zenke, Poole, and Ganguli]{zenke2017img_reg3}
Friedemann Zenke, Ben Poole, and Surya Ganguli.
\newblock Continual learning through synaptic intelligence.
\newblock In \emph{International Conference on Machine Learning}, pages
  3987--3995. PMLR, 2017.

\bibitem[Zhao et~al.(2021{\natexlab{a}})Zhao, Li, Lu, and
  Li]{zhao2021reconstructive}
Bin Zhao, Haopeng Li, Xiaoqiang Lu, and Xuelong Li.
\newblock Reconstructive sequence-graph network for video summarization.
\newblock \emph{IEEE Transactions on Pattern Analysis and Machine
  Intelligence}, 44\penalty0 (5):\penalty0 2793--2801, 2021{\natexlab{a}}.

\bibitem[Zhao et~al.(2021{\natexlab{b}})Zhao, Qin, Su, Fu, Lin, and
  Li]{zhao2021civc}
Hanbin Zhao, Xin Qin, Shihao Su, Yongjian Fu, Zibo Lin, and Xi~Li.
\newblock When video classification meets incremental classes.
\newblock In \emph{Proceedings of the 29th ACM International Conference on
  Multimedia}, pages 880--889, 2021{\natexlab{b}}.

\bibitem[Zhi et~al.(2021)Zhi, Tong, Wang, and Wu]{zhi2021mgsampler}
Yuan Zhi, Zhan Tong, Limin Wang, and Gangshan Wu.
\newblock Mgsampler: An explainable sampling strategy for video action
  recognition.
\newblock In \emph{Proceedings of the IEEE/CVF International Conference on
  Computer Vision}, pages 1513--1522, 2021.

\bibitem[Zhu et~al.(2020)Zhu, Lu, Li, and Zhou]{zhu2020dsnet}
Wencheng Zhu, Jiwen Lu, Jiahao Li, and Jie Zhou.
\newblock Dsnet: A flexible detect-to-summarize network for video
  summarization.
\newblock \emph{IEEE Transactions on Image Processing}, 30:\penalty0 948--962,
  2020.

\end{thebibliography}


\begin{thebibliography}{50}
\providecommand{\natexlab}[1]{#1}
\providecommand{\url}[1]{\texttt{#1}}
\expandafter\ifx\csname urlstyle\endcsname\relax
  \providecommand{\doi}[1]{doi: #1}\else
  \providecommand{\doi}{doi: \begingroup \urlstyle{rm}\Url}\fi

\bibitem[Ali et~al.(2021)Ali, Touvron, Caron, Bojanowski, Douze, Joulin,
  Laptev, Neverova, Synnaeve, Verbeek, et~al.]{xcit}
Alaaeldin Ali, Hugo Touvron, Mathilde Caron, Piotr Bojanowski, Matthijs Douze,
  Armand Joulin, Ivan Laptev, Natalia Neverova, Gabriel Synnaeve, Jakob
  Verbeek, et~al.
\newblock Xcit: Cross-covariance image transformers.
\newblock \emph{NeurIPS}, 2021.

\bibitem[Ba et~al.(2016)Ba, Kiros, and Hinton]{LN}
Jimmy~Lei Ba, Jamie~Ryan Kiros, and Geoffrey~E Hinton.
\newblock Layer normalization.
\newblock \emph{arXiv preprint arXiv:1607.06450}, 2016.

\bibitem[Baevski et~al.(2022)Baevski, Hsu, Xu, Babu, Gu, and Auli]{data2vec}
Alexei Baevski, Wei-Ning Hsu, Qiantong Xu, Arun Babu, Jiatao Gu, and Michael
  Auli.
\newblock Data2vec: A general framework for self-supervised learning in speech,
  vision and language.
\newblock \emph{arXiv preprint arXiv:2202.03555}, 2022.

\bibitem[Bao et~al.(2022)Bao, Dong, Piao, and Wei]{beit}
Hangbo Bao, Li~Dong, Songhao Piao, and Furu Wei.
\newblock Beit: Bert pre-training of image transformers.
\newblock In \emph{ICLR}, 2022.

\bibitem[Bar et~al.(2022)Bar, Wang, Kantorov, Reed, Herzig, Chechik, Rohrbach,
  Darrell, and Globerson]{detreg}
Amir Bar, Xin Wang, Vadim Kantorov, Colorado~J Reed, Roei Herzig, Gal Chechik,
  Anna Rohrbach, Trevor Darrell, and Amir Globerson.
\newblock Detreg: Unsupervised pretraining with region priors for object
  detection.
\newblock In \emph{CVPR}, 2022.

\bibitem[Beal et~al.(2020)Beal, Kim, Tzeng, Park, Zhai, and Kislyuk]{ViT-RCNN}
Josh Beal, Eric Kim, Eric Tzeng, Dong~Huk Park, Andrew Zhai, and Dmitry
  Kislyuk.
\newblock Toward transformer-based object detection.
\newblock \emph{arXiv preprint arXiv:2012.09958}, 2020.

\bibitem[Carion et~al.(2020)Carion, Massa, Synnaeve, Usunier, Kirillov, and
  Zagoruyko]{DETR}
Nicolas Carion, Francisco Massa, Gabriel Synnaeve, Nicolas Usunier, Alexander
  Kirillov, and Sergey Zagoruyko.
\newblock End-to-end object detection with transformers.
\newblock In \emph{ECCV}, 2020.

\bibitem[Chen et~al.(2021)Chen, Du, Yang, Beyer, Zhai, Lin, Chen, Li, Song,
  Wang, et~al.]{uvit}
Wuyang Chen, Xianzhi Du, Fan Yang, Lucas Beyer, Xiaohua Zhai, Tsung-Yi Lin,
  Huizhong Chen, Jing Li, Xiaodan Song, Zhangyang Wang, et~al.
\newblock A simple single-scale vision transformer for object localization and
  instance segmentation.
\newblock \emph{arXiv preprint arXiv:2112.09747}, 2021.

\bibitem[Dai et~al.(2021{\natexlab{a}})Dai, Cai, Lin, and Chen]{Up-detr}
Zhigang Dai, Bolun Cai, Yugeng Lin, and Junying Chen.
\newblock Up-detr: Unsupervised pre-training for object detection with
  transformers.
\newblock In \emph{CVPR}, 2021{\natexlab{a}}.

\bibitem[Dai et~al.(2021{\natexlab{b}})Dai, Liu, Le, and Tan]{coatnet}
Zihang Dai, Hanxiao Liu, Quoc Le, and Mingxing Tan.
\newblock Coatnet: Marrying convolution and attention for all data sizes.
\newblock \emph{NeurIPS}, 2021{\natexlab{b}}.

\bibitem[Devlin et~al.(2019)Devlin, Chang, Lee, and Toutanova]{BERT}
Jacob Devlin, Ming-Wei Chang, Kenton Lee, and Kristina Toutanova.
\newblock Bert: Pre-training of deep bidirectional transformers for language
  understanding.
\newblock \emph{NAACL}, 2019.

\bibitem[Dong et~al.(2021)Dong, Bao, Zhang, Chen, Zhang, Yuan, Chen, Wen, and
  Yu]{peco}
Xiaoyi Dong, Jianmin Bao, Ting Zhang, Dongdong Chen, Weiming Zhang, Lu~Yuan,
  Dong Chen, Fang Wen, and Nenghai Yu.
\newblock Peco: Perceptual codebook for bert pre-training of vision
  transformers.
\newblock \emph{arXiv preprint arXiv:2111.12710}, 2021.

\bibitem[Dosovitskiy et~al.(2021)Dosovitskiy, Beyer, Kolesnikov, Weissenborn,
  Zhai, Unterthiner, Dehghani, Minderer, Heigold, Gelly, Uszkoreit, and
  Houlsby]{ViT}
Alexey Dosovitskiy, Lucas Beyer, Alexander Kolesnikov, Dirk Weissenborn,
  Xiaohua Zhai, Thomas Unterthiner, Mostafa Dehghani, Matthias Minderer, Georg
  Heigold, Sylvain Gelly, Jakob Uszkoreit, and Neil Houlsby.
\newblock An image is worth 16x16 words: Transformers for image recognition at
  scale.
\newblock In \emph{ICLR}, 2021.

\bibitem[Fan et~al.(2021)Fan, Xiong, Mangalam, Li, Yan, Malik, and
  Feichtenhofer]{MViT}
Haoqi Fan, Bo~Xiong, Karttikeya Mangalam, Yanghao Li, Zhicheng Yan, Jitendra
  Malik, and Christoph Feichtenhofer.
\newblock Multiscale vision transformers.
\newblock In \emph{ICCV}, 2021.

\bibitem[Fang et~al.(2021)Fang, Liao, Wang, Fang, Qi, Wu, Niu, and Liu]{YOLOS}
Yuxin Fang, Bencheng Liao, Xinggang Wang, Jiemin Fang, Jiyang Qi, Rui Wu,
  Jianwei Niu, and Wenyu Liu.
\newblock You only look at one sequence: Rethinking transformer in vision
  through object detection.
\newblock \emph{NeurIPS}, 2021.

\bibitem[Fang et~al.(2022)Fang, Dong, Bao, Wang, and Wei]{cim}
Yuxin Fang, Li~Dong, Hangbo Bao, Xinggang Wang, and Furu Wei.
\newblock Corrupted image modeling for self-supervised visual pre-training.
\newblock \emph{arXiv preprint arXiv:2202.03382}, 2022.

\bibitem[Ghiasi et~al.(2021)Ghiasi, Cui, Srinivas, Qian, Lin, Cubuk, Le, and
  Zoph]{lsj}
Golnaz Ghiasi, Yin Cui, Aravind Srinivas, Rui Qian, Tsung-Yi Lin, Ekin~D Cubuk,
  Quoc~V Le, and Barret Zoph.
\newblock Simple copy-paste is a strong data augmentation method for instance
  segmentation.
\newblock In \emph{CVPR}, 2021.

\bibitem[He et~al.(2016)He, Zhang, Ren, and Sun]{ResNet}
Kaiming He, Xiangyu Zhang, Shaoqing Ren, and Jian Sun.
\newblock Deep residual learning for image recognition.
\newblock In \emph{CVPR}, 2016.

\bibitem[He et~al.(2017)He, Gkioxari, Doll{\'a}r, and Girshick]{MaskRCNN}
Kaiming He, Georgia Gkioxari, Piotr Doll{\'a}r, and Ross~B. Girshick.
\newblock Mask r-cnn.
\newblock In \emph{ICCV}, 2017.

\bibitem[He et~al.(2021)He, Chen, Xie, Li, Doll{\'a}r, and Girshick]{mae}
Kaiming He, Xinlei Chen, Saining Xie, Yanghao Li, Piotr Doll{\'a}r, and Ross
  Girshick.
\newblock Masked autoencoders are scalable vision learners.
\newblock \emph{arXiv preprint arXiv:2111.06377}, 2021.

\bibitem[Hendrycks and Gimpel(2016)]{GELU}
Dan Hendrycks and Kevin Gimpel.
\newblock Gaussian error linear units (gelus).
\newblock \emph{arXiv preprint arXiv:1606.08415}, 2016.

\bibitem[Krizhevsky et~al.(2012)Krizhevsky, Sutskever, and Hinton]{AlexNet}
Alex Krizhevsky, Ilya Sutskever, and Geoffrey~E Hinton.
\newblock Imagenet classification with deep convolutional neural networks.
\newblock \emph{NeurIPS}, 2012.

\bibitem[LeCun et~al.(1989)LeCun, Boser, Denker, Henderson, Howard, Hubbard,
  and Jackel]{lecun1989backpropagation}
Yann LeCun, Bernhard Boser, John~S Denker, Donnie Henderson, Richard~E Howard,
  Wayne Hubbard, and Lawrence~D Jackel.
\newblock Backpropagation applied to handwritten zip code recognition.
\newblock \emph{Neural computation}, 1989.

\bibitem[Li et~al.(2021{\natexlab{a}})Li, Wu, Fan, Mangalam, Xiong, Malik, and
  Feichtenhofer]{MViTv2}
Yanghao Li, Chao-Yuan Wu, Haoqi Fan, Karttikeya Mangalam, Bo~Xiong, Jitendra
  Malik, and Christoph Feichtenhofer.
\newblock Improved multiscale vision transformers for classification and
  detection.
\newblock \emph{arXiv preprint arXiv:2112.01526}, 2021{\natexlab{a}}.

\bibitem[Li et~al.(2021{\natexlab{b}})Li, Xie, Chen, Dollar, He, and
  Girshick]{benchmarking}
Yanghao Li, Saining Xie, Xinlei Chen, Piotr Dollar, Kaiming He, and Ross
  Girshick.
\newblock Benchmarking detection transfer learning with vision transformers.
\newblock \emph{arXiv preprint arXiv:2111.11429}, 2021{\natexlab{b}}.

\bibitem[Lin et~al.(2014)Lin, Maire, Belongie, Hays, Perona, Ramanan,
  Doll{\'a}r, and Zitnick]{COCO}
Tsung-Yi Lin, Michael Maire, Serge Belongie, James Hays, Pietro Perona, Deva
  Ramanan, Piotr Doll{\'a}r, and C~Lawrence Zitnick.
\newblock Microsoft coco: Common objects in context.
\newblock In \emph{ECCV}, 2014.

\bibitem[Lin et~al.(2017{\natexlab{a}})Lin, Doll{\'a}r, Girshick, He,
  Hariharan, and Belongie]{FPN}
Tsung-Yi Lin, Piotr Doll{\'a}r, Ross Girshick, Kaiming He, Bharath Hariharan,
  and Serge Belongie.
\newblock Feature pyramid networks for object detection.
\newblock In \emph{CVPR}, 2017{\natexlab{a}}.

\bibitem[Lin et~al.(2017{\natexlab{b}})Lin, Goyal, Girshick, He, and
  Doll{\'a}r]{RetinaNet}
Tsung-Yi Lin, Priya Goyal, Ross Girshick, Kaiming He, and Piotr Doll{\'a}r.
\newblock Focal loss for dense object detection.
\newblock In \emph{ICCV}, 2017{\natexlab{b}}.

\bibitem[Liu et~al.(2021)Liu, Lin, Cao, Hu, Wei, Zhang, Lin, and Guo]{Swin}
Ze~Liu, Yutong Lin, Yue Cao, Han Hu, Yixuan Wei, Zheng Zhang, Stephen Lin, and
  Baining Guo.
\newblock Swin transformer: Hierarchical vision transformer using shifted
  windows.
\newblock In \emph{ICCV}, 2021.

\bibitem[Odena et~al.(2016)Odena, Dumoulin, and Olah]{artifacts}
Augustus Odena, Vincent Dumoulin, and Chris Olah.
\newblock Deconvolution and checkerboard artifacts.
\newblock \emph{Distill}, 2016.
\newblock URL \url{http://distill.pub/2016/deconv-checkerboard}.

\bibitem[Radford et~al.(2018)Radford, Narasimhan, Salimans, and Sutskever]{GPT}
Alec Radford, Karthik Narasimhan, Tim Salimans, and Ilya Sutskever.
\newblock Improving language understanding by generative pre-training.
\newblock 2018.

\bibitem[Radford et~al.(2021)Radford, Kim, Hallacy, Ramesh, Goh, Agarwal,
  Sastry, Askell, Mishkin, Clark, et~al.]{CLIP}
Alec Radford, Jong~Wook Kim, Chris Hallacy, Aditya Ramesh, Gabriel Goh,
  Sandhini Agarwal, Girish Sastry, Amanda Askell, Pamela Mishkin, Jack Clark,
  et~al.
\newblock Learning transferable visual models from natural language
  supervision.
\newblock In \emph{ICML}, 2021.

\bibitem[Raffel et~al.(2020)Raffel, Shazeer, Roberts, Lee, Narang, Matena,
  Zhou, Li, and Liu]{rel_pos}
Colin Raffel, Noam Shazeer, Adam Roberts, Katherine Lee, Sharan Narang, Michael
  Matena, Yanqi Zhou, Wei Li, and Peter~J. Liu.
\newblock Exploring the limits of transfer learning with a unified text-to-text
  transformer.
\newblock \emph{JMLR}, 2020.

\bibitem[Ren et~al.(2015)Ren, He, Girshick, and Sun]{Faster-R-CNN}
Shaoqing Ren, Kaiming He, Ross Girshick, and Jian Sun.
\newblock Faster r-cnn: Towards real-time object detection with region proposal
  networks.
\newblock \emph{NeurIPS}, 2015.

\bibitem[Russakovsky et~al.(2015)Russakovsky, Deng, Su, Krause, Satheesh, Ma,
  Huang, Karpathy, Khosla, Bernstein, et~al.]{ImageNet-1k}
Olga Russakovsky, Jia Deng, Hao Su, Jonathan Krause, Sanjeev Satheesh, Sean Ma,
  Zhiheng Huang, Andrej Karpathy, Aditya Khosla, Michael Bernstein, et~al.
\newblock Imagenet large scale visual recognition challenge.
\newblock \emph{IJCV}, 2015.

\bibitem[Simonyan and Zisserman(2015)]{VGG}
Karen Simonyan and Andrew Zisserman.
\newblock Very deep convolutional networks for large-scale image recognition.
\newblock In \emph{ICLR}, 2015.

\bibitem[Tan and Le(2019)]{EfficientNet}
Mingxing Tan and Quoc Le.
\newblock Efficientnet: Rethinking model scaling for convolutional neural
  networks.
\newblock In \emph{ICML}, 2019.

\bibitem[Tan et~al.(2020)Tan, Pang, and Le]{EfficientDet}
Mingxing Tan, Ruoming Pang, and Quoc~V Le.
\newblock Efficientdet: Scalable and efficient object detection.
\newblock In \emph{CVPR}, 2020.

\bibitem[Tian et~al.(2019)Tian, Shen, Chen, and He]{fcos}
Zhi Tian, Chunhua Shen, Hao Chen, and Tong He.
\newblock Fcos: Fully convolutional one-stage object detection.
\newblock In \emph{ICCV}, 2019.

\bibitem[Vaswani et~al.(2017)Vaswani, Shazeer, Parmar, Uszkoreit, Jones, Gomez,
  Kaiser, and Polosukhin]{Transformer}
Ashish Vaswani, Noam Shazeer, Niki Parmar, Jakob Uszkoreit, Llion Jones,
  Aidan~N Gomez, {\L}ukasz Kaiser, and Illia Polosukhin.
\newblock Attention is all you need.
\newblock \emph{NeurIPS}, 2017.

\bibitem[Vaswani et~al.(2021)Vaswani, Ramachandran, Srinivas, Parmar, Hechtman,
  and Shlens]{HaloNet}
Ashish Vaswani, Prajit Ramachandran, Aravind Srinivas, Niki Parmar, Blake
  Hechtman, and Jonathon Shlens.
\newblock Scaling local self-attention for parameter efficient visual
  backbones.
\newblock In \emph{CVPR}, 2021.

\bibitem[Wang et~al.(2021)Wang, Xie, Li, Fan, Song, Liang, Lu, Luo, and
  Shao]{PVT}
Wenhai Wang, Enze Xie, Xiang Li, Deng-Ping Fan, Kaitao Song, Ding Liang, Tong
  Lu, Ping Luo, and Ling Shao.
\newblock Pyramid vision transformer: A versatile backbone for dense prediction
  without convolutions.
\newblock In \emph{ICCV}, 2021.

\bibitem[Wei et~al.(2021)Wei, Fan, Xie, Wu, Yuille, and
  Feichtenhofer]{maskfeat}
Chen Wei, Haoqi Fan, Saining Xie, Chao-Yuan Wu, Alan Yuille, and Christoph
  Feichtenhofer.
\newblock Masked feature prediction for self-supervised visual pre-training.
\newblock \emph{arXiv preprint arXiv:2112.09133}, 2021.

\bibitem[Wu et~al.(2019)Wu, Kirillov, Massa, Lo, and Girshick]{detectron2}
Yuxin Wu, Alexander Kirillov, Francisco Massa, Wan-Yen Lo, and Ross Girshick.
\newblock Detectron2.
\newblock \url{https://github.com/facebookresearch/detectron2}, 2019.

\bibitem[Xiao et~al.(2021)Xiao, Dollar, Singh, Mintun, Darrell, and
  Girshick]{xiao2021early}
Tete Xiao, Piotr Dollar, Mannat Singh, Eric Mintun, Trevor Darrell, and Ross
  Girshick.
\newblock Early convolutions help transformers see better.
\newblock In \emph{NeurIPS}, 2021.

\bibitem[Xie et~al.(2017)Xie, Girshick, Doll{\'a}r, Tu, and He]{resnext}
Saining Xie, Ross Girshick, Piotr Doll{\'a}r, Zhuowen Tu, and Kaiming He.
\newblock Aggregated residual transformations for deep neural networks.
\newblock In \emph{CVPR}, 2017.

\bibitem[Xie et~al.(2022)Xie, Zhang, Cao, Lin, Bao, Yao, Dai, and Hu]{simmim}
Zhenda Xie, Zheng Zhang, Yue Cao, Yutong Lin, Jianmin Bao, Zhuliang Yao,
  Qi~Dai, and Han Hu.
\newblock Simmim: A simple framework for masked image modeling.
\newblock In \emph{CVPR}, 2022.

\bibitem[Xu et~al.(2021)Xu, Xu, Chang, and Tu]{CoaT}
Weijian Xu, Yifan Xu, Tyler Chang, and Zhuowen Tu.
\newblock Co-scale conv-attentional image transformers.
\newblock In \emph{ICCV}, 2021.

\bibitem[Zhou et~al.(2022)Zhou, Wei, Wang, Shen, Xie, Yuille, and Kong]{ibot}
Jinghao Zhou, Chen Wei, Huiyu Wang, Wei Shen, Cihang Xie, Alan Yuille, and Tao
  Kong.
\newblock ibot: Image bert pre-training with online tokenizer.
\newblock In \emph{ICLR}, 2022.

\bibitem[Zhu et~al.(2021)Zhu, Su, Lu, Li, Wang, and Dai]{Deformable-DETR}
Xizhou Zhu, Weijie Su, Lewei Lu, Bin Li, Xiaogang Wang, and Jifeng Dai.
\newblock Deformable detr: Deformable transformers for end-to-end object
  detection.
\newblock In \emph{ICLR}, 2021.

\end{thebibliography}

\section*{Checklist}

The checklist follows the references.  Please
read the checklist guidelines carefully for information on how to answer these
questions.  For each question, change the default \answerTODO{} to \answerYes{},
\answerNo{}, or \answerNA{}.  You are strongly encouraged to include a {\bf
justification to your answer}, either by referencing the appropriate section of
your paper or providing a brief inline description.  For example:
\begin{itemize}
  \item Did you include the license to the code and datasets? \answerNA{}
\end{itemize}
Please do not modify the questions and only use the provided macros for your
answers.  Note that the Checklist section does not count towards the page
limit.  In your paper, please delete this instructions block and only keep the
Checklist section heading above along with the questions/answers below.

\begin{enumerate}

\item For all authors...
\begin{enumerate}
  \item Do the main claims made in the abstract and introduction accurately reflect the paper's contributions and scope?
    \answerYes{}
  \item Did you describe the limitations of your work?
    \answerYes{}
  \item Did you discuss any potential negative societal impacts of your work?
    \answerYes{}
  \item Have you read the ethics review guidelines and ensured that your paper conforms to them?
    \answerYes{}
\end{enumerate}

\item If you are including theoretical results...
\begin{enumerate}
  \item Did you state the full set of assumptions of all theoretical results?
    \answerNA{}
        \item Did you include complete proofs of all theoretical results?
    \answerNA{}
\end{enumerate}

\item If you ran experiments...
\begin{enumerate}
  \item Did you include the code, data, and instructions needed to reproduce the main experimental results (either in the supplemental material or as a URL)?
    \answerNo{}
  \item Did you specify all the training details (e.g., data splits, hyperparameters, how they were chosen)?
    \answerYes{}
        \item Did you report error bars (e.g., with respect to the random seed after running experiments multiple times)?
    \answerNo{}
        \item Did you include the total amount of compute and the type of resources used (e.g., type of GPUs, internal cluster, or cloud provider)?
    \answerYes{}
\end{enumerate}

\item If you are using existing assets (e.g., code, data, models) or curating/releasing new assets...
\begin{enumerate}
  \item If your work uses existing assets, did you cite the creators?
    \answerYes{}
  \item Did you mention the license of the assets?
    \answerNo{}
  \item Did you include any new assets either in the supplemental material or as a URL?
    \answerNo{}
  \item Did you discuss whether and how consent was obtained from people whose data you're using/curating?
    \answerNo{}
  \item Did you discuss whether the data you are using/curating contains personally identifiable information or offensive content?
    \answerNo{}
\end{enumerate}

\item If you used crowdsourcing or conducted research with human subjects...
\begin{enumerate}
  \item Did you include the full text of instructions given to participants and screenshots, if applicable?
    \answerNA{}
  \item Did you describe any potential participant risks, with links to Institutional Review Board (IRB) approvals, if applicable?
    \answerNA{}
  \item Did you include the estimated hourly wage paid to participants and the total amount spent on participant compensation?
    \answerNA{}
\end{enumerate}

\end{enumerate}


\newpage
\appendix
\definecolor{grey}{RGB}{232,232,232}
\definecolor{baselinecolor}{gray}{.9}
\begin{appendix}
\section{Additional Experiments for Different Memory Budgets}

This is supplementary to Section 4.2 "Comparison with State-of-the-art Results". Limited by space, we only summarize the results of our approach using the same exemplar memory size per class as others and the experiment results of "Memory budget comparison" on UCF101 with 10 steps in the main body. Table~\ref{table:memory_ucf_hmdb} supplements the average accuracy of FrameMaker with different memory budgets under all benchmarks of UCF101 and HMDB51 datasets, which is compared with the best performance of TCD~\cite{park2021tcd}. 

\begin{table}[htbp]
  \caption{Comparison with the state-of-the-art approaches over class-incremental action recognition performance on UCF101 and HMDB51 with different memory budgets. }
  \label{table:memory_ucf_hmdb}
  \tablestyle{5pt}{1.2}
  \centering
  \begin{tabular}{ccccccccccc}
    \toprule[1.15pt]
    ~ &  \multicolumn{6}{c}{UCF101} & \multicolumn{4}{c}{HMDB51}   \\
    \cmidrule(r){1-1}\cmidrule(r){2-7}\cmidrule(r){8-11}
    Num. of Classes  & \multicolumn{2}{c}{$10\times5$ stages}   & \multicolumn{2}{c}{$5\times10$ stages}   & \multicolumn{2}{c}{$2\times25$ stages}  & \multicolumn{2}{c}{$5\times5$ stages}  & \multicolumn{2}{c}{$1\times25$ stages}\\
    Memory Per Class  & CNN & NME & CNN & NME & CNN & NME & CNN & NME & CNN & NME   \\
    \cmidrule(r){1-1}\cmidrule(r){2-3}\cmidrule(r){4-5}\cmidrule(r){6-7}\cmidrule(r){8-9}\cmidrule(r){10-11}
    $8\text{F}\times5\text{V}$=6.0Mb~\cite{park2021tcd} & 74.89 & 77.16 & 73.43 & 75.35 & 72.19 & 74.01 & 45.34 & 50.36 & 40.47 & 46.66      \\
    \cmidrule(r){1-1}\cmidrule(r){2-3}\cmidrule(r){4-5}\cmidrule(r){6-7}\cmidrule(r){8-9}\cmidrule(r){10-11}
    $1\text{F}\times1\text{V}$=0.15Mb & 58.75 & 73.97 & 49.37 & 70.78 & 48.31 & 65.86 & 36.08 & 43.82 & 32.53 & 39.00     \\
    $1\text{F}\times2\text{V}$=0.3Mb & 67.59 & 76.18 & 62.06 & 74.18 & 55.84 & 72.88 & 39.33 & 46.07 & 34.05 & 43.18     \\
    $1\text{F}\times5\text{V}$=0.75Mb & 73.05 & 76.70 & 72.93 & 76.64 & 68.79 & 75.36 & 43.39 & 46.88 & 38.95 & 44.18     \\
    $1\text{F}\times8\text{V}$=1.2Mb & 76.09 & 77.88 & 73.64 & 76.98 & 73.57 & 76.76 & 46.25 & 49.02 & 40.80 & 46.97     \\
    $1\text{F}\times16\text{V}$=2.4Mb & 77.24 & 78.06 & 75.19 & 77.43 & 74.92 & 76.92 & 46.97 & 50.42 & 41.96 & 47.03\\
    $1\text{F}\times40\text{V}$=6.0Mb & \textbf{78.13} & \textbf{78.64} & \textbf{76.38} & \textbf{78.14} & \textbf{75.77} & \textbf{77.49} & \textbf{47.54} & \textbf{51.12} & \textbf{42.65} & \textbf{47.37}     \\
    \bottomrule[1.15pt]
  \end{tabular}
\end{table}

As can be seen from Table~\ref{table:memory_ucf_hmdb}, under other settings, the results of our method are similar to that on UCF101 with 10 steps. When using about 20\% of the storage space, our method can achieve better performance than TCD, and the accuracy is further improved when the number of saved exemplars is increased, which also proves the robustness of our approach. 

\section{Additional Analysis of the Balance Weights}
This is supplementary to Section 4.3 "Ablation Study". We further discuss the sensitivity to the balance weight for each term in $L_{\text{fc}}$ on UCF101 with 10 steps. 

We first discuss the sensitivity of ${\alpha}$ and ${\beta}$, the balance weights for ${L^\text{c}_{\text{f}}}$ and $L^\text{c}_{\text{ce}}$ which are used to achieve stronger constraint on $W_i^k$ for its effective training. Table~\ref{table:sensitivity_ab} shows the performance of frame condensing (without instance-specific prompt) under various combinations of ${\alpha}$ and ${\beta}$. We find the performance under the combination of $\{{\alpha}=1, {\beta}=1\}$ consistently exceeds others. Although a slightly higher performance is obtained on NME under $\{{\alpha}=2, {\beta}=1\}$, it is difficult to achieve the same gain on the two different metrics after continuous adjustment. 

\begin{table}[htbp]
\centering
\caption{Sensitivity of the performance of Frame Condensing to ${\alpha}$ and ${\beta}$ on UCF101 with 10 steps. Default settings are marked in \colorbox{baselinecolor}{gray}.}
\label{table:sensitivity_ab}
\tablestyle{6.2pt}{1.2}
\begin{tabular}{cccccccccccc}
    \toprule[1.15pt]
    ${\alpha}({L^\text{c}_{\text{f}})}$ & 1 & 2 & 5 & 0.5 & 0.1 & 0.01 & 1 & 1 & 1 & 1 & 1 \\
    ${\beta}(L^\text{c}_{\text{ce}})$ & 1 & 1 & 1 & 1 & 1 & 1 & 2 & 5 & 0.5 & 0.1 & 0.01 \\
    \cmidrule(r){1-1}\cmidrule(r){2-12}
    CNN & \cellcolor{grey}\textbf{72.29} & 72.27 & 71.83 & 72.22 & 72.07 & 71.94 & 71.93 & 71.44 & 72.14 & 71.96 & 71.92 \\
    NME & \cellcolor{grey}76.42 & \textbf{76.43} & 76.36 & 76.25 & 76.21 & 76.13 & 76.37 & 76.23 & 76.22 & 75.91 & 75.84\\
    
    \bottomrule[1.15pt]
\end{tabular}
\end{table}

Then we further analyze sensitivity of ${\gamma}$ and ${\eta}$, the balance weights for ${L^\text{p}_{\text{f}}}$ and $L^\text{p}_{\text{ce}}$, when setting ${\alpha}$ and ${\beta}$ as the optimal value, $1$. Table~\ref{table:sensitivity_cd} shows the performance of FrameMaker (frame condensing with instance-specific prompt) under various combinations of ${\gamma}$ and ${\eta}$. 

\begin{table}[htbp]
\centering
\caption{Sensitivity of the performance of FrameMaker to ${\gamma}$ and ${\eta}$ $({\alpha}=1, {\beta}=1)$ on UCF101 with 10 steps. Default settings are marked in \colorbox{baselinecolor}{gray}.}
\label{table:sensitivity_cd}
\tablestyle{6.2pt}{1.2}
\begin{tabular}{ccccccccccccc}
    \toprule[1.15pt]
    ${\gamma}({L^\text{p}_{\text{f}}})$ & 1 & 2 & 5 & 0.5 & 0.1 & 0.01 & 1 & 1 & 1 & 1 & 1 \\
    ${\eta}(L^\text{p}_{\text{ce}})$ & 1 & 1 & 1 & 1 & 1 & 1 & 2 & 5 & 0.5 & 0.1 & 0.01 \\
    \cmidrule(r){1-1}\cmidrule(r){2-12}
    CNN & \cellcolor{grey}72.93 & \textbf{72.94} & 72.92 & 72.88 & 72.64 & 72.67 & 72.90 & 72.85 & 72.92 & 72.80 & 72.73 \\
    NME & \cellcolor{grey}76.64 & 76.62 & 76.58 & 76.61 & 76.59 & 76.61 & \textbf{76.65} & 76.58 & 76.61 & 76.55 & 76.51 \\
    \bottomrule[1.15pt]
\end{tabular}
\end{table}

We find the performance under the combination of $\{{\gamma}=1, {\eta}=1\}$ always exceeds others. Some groups of weights make one of the test metrics higher but only $\{{\gamma}=1, {\eta}=1\}$ obtain the best balance of them. Therefore, we set all the balance weights as $1$ finally in our experiments.

\section{Implementation Details}
\begin{figure}[htbp]
  \centering
  \includegraphics[width=1.0\linewidth]{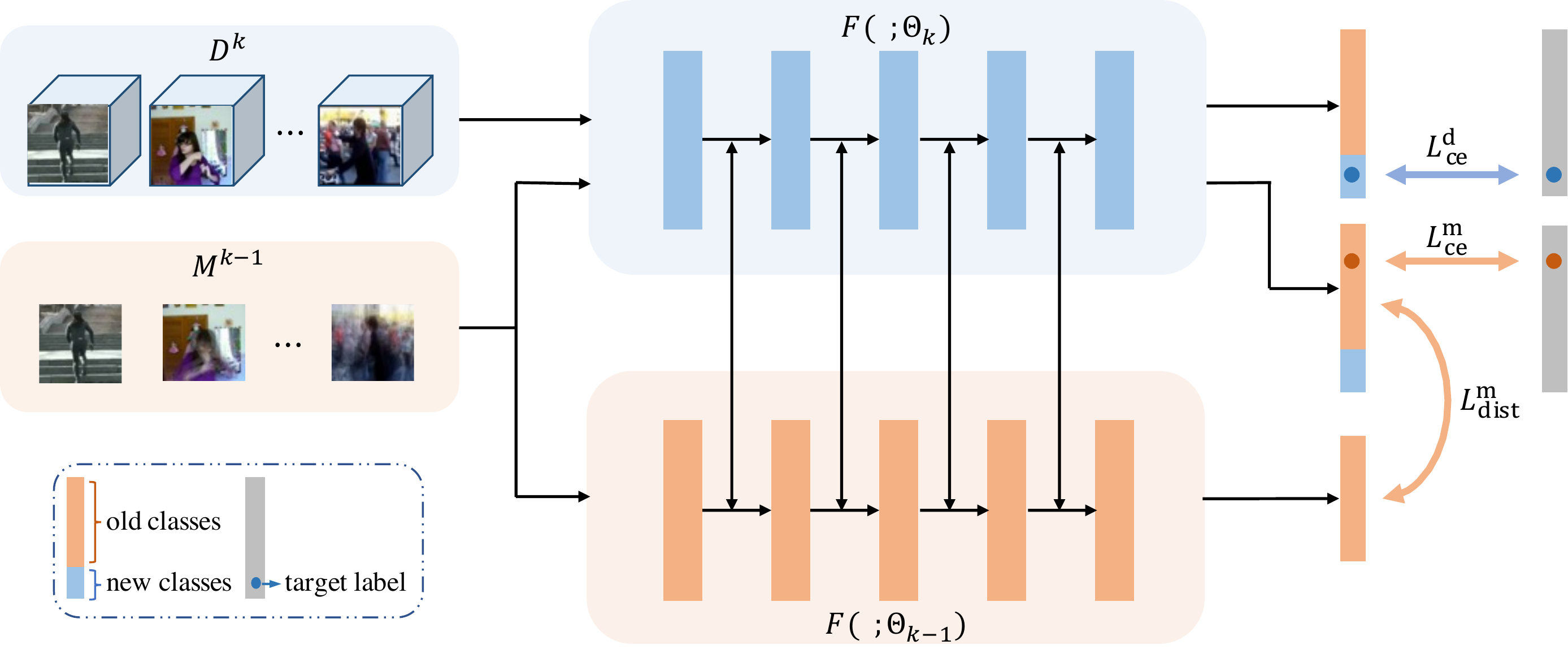}
  \caption{An overview of the training process}
  \label{fig:training process}
\end{figure}

This section provides some additional details about the experiments in the main body.
\paragraph{Training Details.}
This is supplementary to Section 3.3 "Training". We further supplement the elaborate training process and knowledge distillation loss in this section. Figure~\ref{fig:training process} shows the overall framework of training process. 

In the training process of the incremental step $k$, we use the dataset ${D'}^k=D^k\bigcup{M^{k-1}}$ to update the model from $F(;\Theta_{k-1})$,
where $D^k$ is the dataset of the task $k$ which consists of videos belonging to current classes, and ${M^{k-1}}$ is memory bank which contains condensed frames generated by FrameMaker after the step $k-1$. 
The model classifies the samples from the two datasets respectively and use exemplars in old classes for knowledge distillation. 

For the classification task, we adopt cross entropy loss to supervise.  $L^\text{d}_{{\text{ce}}}$ and $L^\text{m}_{{\text{ce}}}$ are the cross entropy losses for video example $V_i^k$ in $D^k$ and condensed frame exemplar $I_i^{k-1}$ in ${M^{k-1}}$. 

\begin{equation}
    L^\text{d}_{{\text{ce}}}={\text{CrossEntropy}(F(V_i^k;\Theta_k),y_i^k)} \ ,
\end{equation}
\begin{equation}
    L^\text{m}_{{\text{ce}}}={\text{CrossEntropy}(F(I_i^{k-1};\Theta_k),y_i^{k-1})}  \ ,
\end{equation}
where $y_i^k$ and $y_i^{k-1}$ is the ground truth label of $V_i^k$ and $I_i^{k-1}$, respectively. 

In order to save the information of old classes better, the knowledge distillation function of PODNet~\cite{douillard2020podnet} is employed on the old classes data ${M^{k-1}}$ and it consists of two components, the spatial distillation function $L_\text{spatial}$ and the final embedding distillation function $L_\text{flat}$. 
\begin{equation}
    \begin{split}
    L_{\text{spatial}}&{(f_l(I_i^{k-1};\Theta_k),f_l(I_i^{k-1};\Theta_{k-1}))} \\
    & =\sum_{c=1}^C \sum_{h=1}^H {\mid\mid \sum_{w=1}^W f_{l,c,w,h}(I_i^{k-1};\Theta_k) - \sum_{w=1}^W f_{l,c,w,h}(I_i^{k-1};\Theta_{k-1}) \mid\mid^2}  \\
    & + \sum_{c=1}^C \sum_{w=1}^W {\mid\mid \sum_{h=1}^H f_{l,c,w,h}(I_i^{k-1};\Theta_k) - \sum_{h=1}^H f_{l,c,w,h}(I_i^{k-1};\Theta_{k-1}) \mid\mid^2}  \ ,
    \end{split}
\end{equation}
where $f_l(;\Theta_k)$ is the output of intermediate convolution layer $l$ of model $F(;\Theta_k)$. C, H, W are the size of channel, height and width of the output feature of layer $l$. 
\begin{equation}
    L_{\text{flat}}{(f(I_i^{k-1};\Theta_k),f(I_i^{k-1};\Theta_{k-1}))}=\mid\mid f(I_i^{k-1};\Theta_k)-f(I_i^{k-1};\Theta_{k-1}) \mid\mid^2  \ ,
\end{equation}
where $f(;\Theta_k)$ is the feature extractor of model $F(;\Theta_k)$. 

The final distillation loss of the condensed frame exemplars is given by: 
\begin{equation}
    L^\text{m}_{{\text{dist}}}= L_{\text{spatial}}{(f_l(I_i^{k-1};\Theta_k),f_l(I_i^{k-1};\Theta_{k-1}))}+L_{\text{flat}}{(f(I_i^{k-1};\Theta_k),f(I_i^{k-1};\Theta_{k-1}))}\ ,
\end{equation}

To simplify, the hyperparameters used in PODNet~\cite{douillard2020podnet} final distillation loss is ignored. 

The total objective function for task $k$ is given by:
\begin{equation}
    L_{\text{cil}}={L^\text{d}_{{\text{ce}}} + L^\text{m}_{{\text{ce}}} + L^\text{m}_{{\text{dist}}}} \ ,
\end{equation}
which is the same as Eq. 8 in the manuscript. 

\paragraph{Dataset Details.}
Following TCD~\cite{park2021tcd}, we use three action recognition datasets, UCF101~\cite{soomro2012ucf101}, HMDB51~\cite{kuehne2011hmdb} and Something-Something V2~\cite{goyal2017something}. 
The HMDB51 dataset is a large collection of realistic videos from various sources which can be found in \href{https://serre-lab.clps.brown.edu/resource/hmdb-a-large-human-motion-database/}{https://serre-lab.clps.brown.edu/resource/hmdb-a-large-human-motion-database/}. UCF101 dataset is an extension of UCF50~\cite{reddy2013recognizing} and consists of 13,320 video clips, which can be found in \href{https://www.crcv.ucf.edu/data/UCF101.php}{https://www.crcv.ucf.edu/data/UCF101.php}.  
Something-SomethingV2 dataset is a large-scale motion-sensitive dataset that shows humans performing pre-defined basic actions with everyday objects, which can be found in \href{https://developer.qualcomm.com/software/ai-datasets/something-something}{https://developer.qualcomm.com/software/ai-datasets/something-something}. 
For UCF101, we have three settings, $51+10\times5$ stages, $51+5\times10$ stages and $51+1\times25$ stages. For HMDB51, the number of initial task classes is 26, and the remaining 25 classes are separated into 5 or 25 groups. For Something-Something V2, the benchmarks are $84+10\times9$ stages and $84+5\times18$ stages. The random seed we use is also the same as TCD~\cite{park2021tcd}. 

\paragraph{Model and Hyperparameter Setup.}
TSM~\cite{lin2019tsm} is employed as our backbone, and we follow the data preprocessing procedure and the basic training setting of TSM. We train a ResNet-34 TSM for UCF101 and a ResNet-50 TSM for HMDB51 and Something-Something V2. Each incremental training procedure takes 50 epochs. We use SGD optimizer with an initial learning rate of 0.04 for UCF101 and Something-Something V2, and 0.002 for HMDB51, which is reduced by half after 25 epochs. For incremental steps, we set the learning rate for all datasets as 0.001 and make the same reduction. The learning rate is set as 0.01 and 0.001 for condensing weights and the instance-specific prompt, respectively, and the total iteration for their optimization is set to be 8k. 
\paragraph{Hardware Information.}
We train all models on eight NVIDIA V100 GPUs and use PyTorch~\cite{paszke2019pytorch} for all our experiments.

\section{Visualizations}

\begin{figure}
  \centering
  \includegraphics[width=1.0\linewidth]{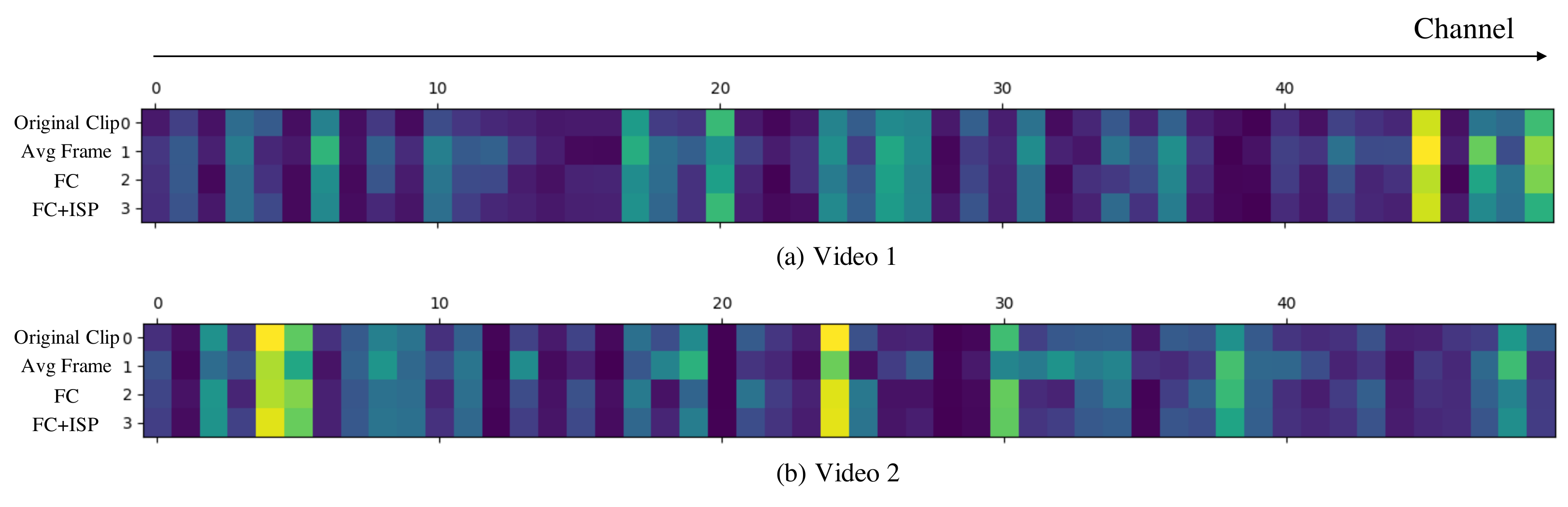}
  \caption{
  Channel comparisons between different inputs on HMDB51 . Each line indicates the feature channel activation value obtained with the specified frame as input. FC and ISP are abbreviations of our proposed Frame Condensing and Instance-Specific Prompt, respectively.}
  \label{fig:prompt_sim}
\end{figure}

\begin{figure}
  \centering
  \includegraphics[width=1.0\linewidth]{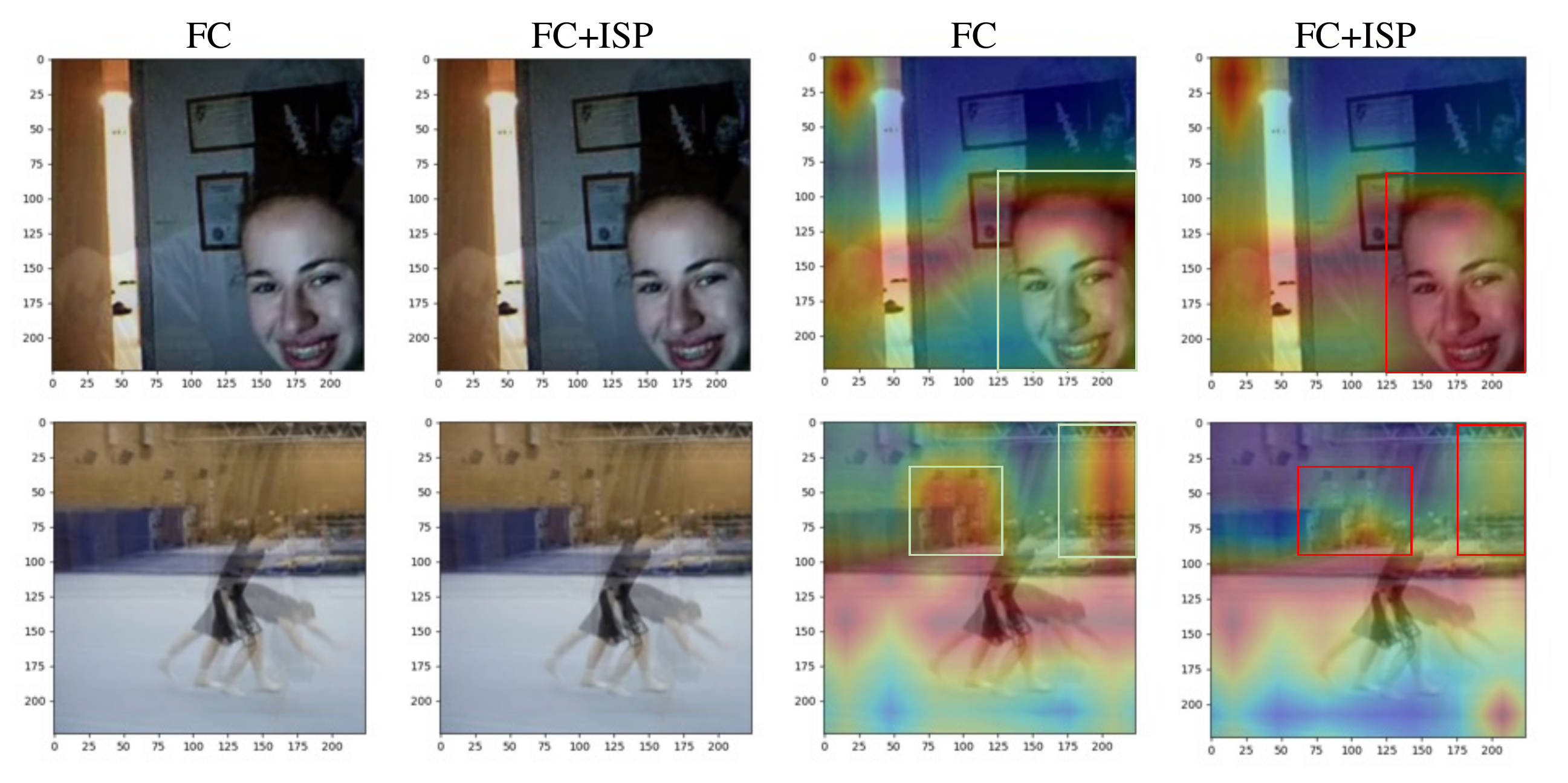}
  \caption{Visualizations of condensed frames and GradCAM maps on HMDB51. FC and ISP refer to proposed Frame Condensing and Instance-Specific Prompt, respectively.}
  \label{fig:cam_vis}
\end{figure}

\textbf{Visual comparison of features.} Figure~\ref{fig:prompt_sim} plots visual comparisons of the  extracted features with different frames as model inputs. We can summarize that the average frame shows the largest feature difference compared to the features extracted from the original clips. Our proposed Frame Condensing and Instance-Specific Prompt can effectively reduce the feature differences with the original clips and thus effectively improve the feature quality of condensed frames.

\textbf{Visualizations of condensed frames.} Figure~\ref{fig:cam_vis} plots the condensed frames, the sum of the condensed frames and their prompts, and the activation region generated by the GradCAM technology.
From the results, we observe the learned prompts have very small magnitudes (aver-
age value is 0.03), hence no significant difference is observed actually.
However, from the GradCAM attention maps, we can observe the condensed frames without prompts may focus on background regions, while the learned prompts tend to increase the attention of motion regions or reduce the attention of the backgrounds.
We think the reason behind this is that the collapsed temporal dynamics may easily lead the model pay more attention to the background to some extent, while the prompt can guide the model to refocus on the motion region and weaken the interference from backgrounds.

\section{Training Prompt without Condensed Frame}
\begin{figure}
  \centering
  \includegraphics[width=1.0\linewidth]{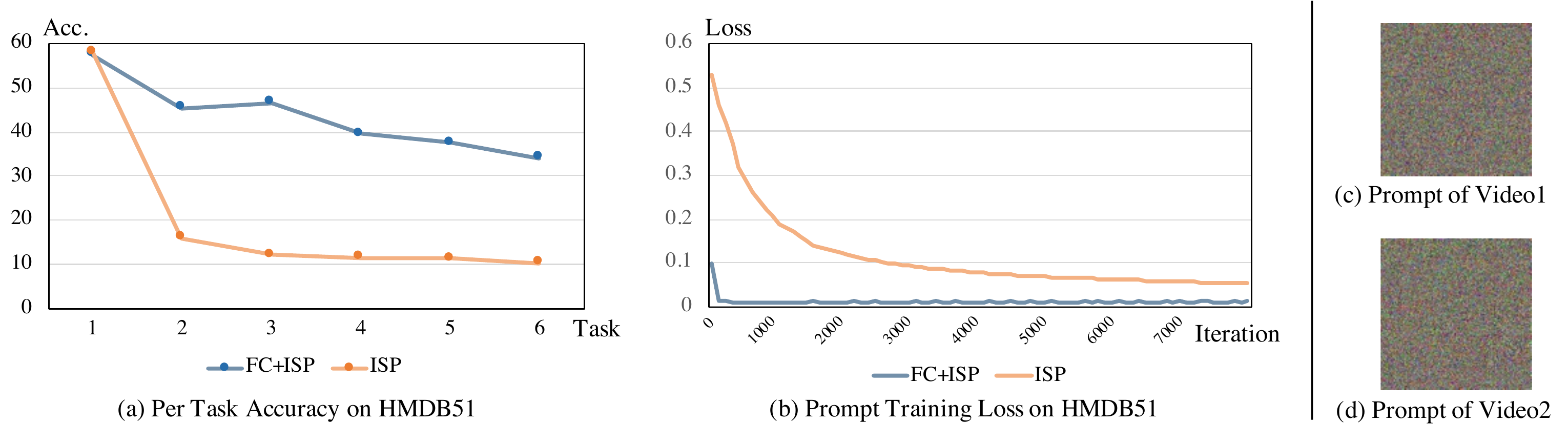}
  \caption{Training prompts without condensed frames. (a): Per-task accuracy on HMDB51. The performance  of Instance-Specific Prompt without condensed frames drops quickly. (b): The training loss of prompts on HMDB51. Instance-Specific Prompt without condensed frames is hard to optimize.  (c) and (d): Learned prompts of two videos.  There is no intuitive semantic information in prompt.}
  \label{fig:isp_without_fc}
\end{figure}

We attempt to train Instance-Specific Prompt with out condensed frames, and the learned prompting parameters are visualized in Figure~\ref{fig:isp_without_fc}(c) and (d). We observe that the learned prompts have no intuitive semantics. To further understand this phenomenon, we plot the per-task accuracy and the loss curve of the prompts in Figure~\ref{fig:isp_without_fc}(a) and (b), respectively. We can summarize two points: (i) the learned prompts without condensed frames cannot fight against catastrophic forgetting, which indicates that the semantic information of the learned prompts is weak; (ii) the loss curve of the prompts without condensed frames is much higher than that with condensed frames, which implies that it is difficult to train a single prompts from scratch. Thus, it is crucial that the condensed frames provide a good initialization for the prompts to avoid the local minima of prompts.
\end{appendix}


\end{document}